\documentclass[10pt,twocolumn,letterpaper]{article}

\usepackage[pagenumbers]{cvpr} 

\definecolor{cvprblue}{rgb}{0.21,0.49,0.74}
\usepackage[pagebackref,breaklinks,colorlinks,allcolors=cvprblue]{hyperref}

\usepackage{makecell,multirow, graphicx}
\usepackage{rotating}
\usepackage{arydshln}
\usepackage{pifont}
\usepackage{arydshln}
\usepackage{graphicx}
\usepackage{subcaption}
\usepackage{framed}
\usepackage{makecell}

\let\tb=\textbf
\let\mb=\mathbf
\let\tu=\underline

\definecolor{green}{RGB}{0, 153, 76}
\definecolor{red}{RGB}{204, 0, 0}
\definecolor{dark_grey}{RGB}{41, 70, 91}

\newcommand{\cmark}{{\color{green}\checkmark}}
\newcommand{\xmark}{{\color{red}\ding{55}}}
\newcommand{\dgt}[1]{\textcolor{dark_grey}{#1}}

\def\paperID{17042} 
\def\confName{CVPR}
\def\confYear{2025}

\newcommand{\sstar}{\textsuperscript{*}}
\newcommand{\sdag}{\textsuperscript{\dag}}

\title{Accurate Scene Text Recognition with Efficient Model Scaling and Cloze Self-Distillation}

\author{Andrea Maracani$^{1}$\thanks{Main Authorship. \texttt{a.maracani@samsung.com}} \quad Savas Ozkan$^{1}$\thanks{Senior Authorship} \quad Sijun Cho$^{2}$ \quad Hyowon Kim$^{2}$ \quad Eunchung Noh$^{2}$ \\ Jeongwon Min$^{2}$ \quad Cho Jung Min $^{2}$ \quad Dookun Park$^{2}$ \quad Mete Ozay$^{1}$ \\
$^{1}$Samsung R\&D Institute UK \quad $^{2}$Samsung Electronics \\
}

\begin{document}
\maketitle
\begin{abstract}
Scaling architectures have been proven effective for improving Scene Text Recognition (STR), but the individual contribution of vision encoder and text decoder scaling remain under-explored. In this work, we present an in-depth empirical analysis and demonstrate that, contrary to previous observations, scaling the decoder yields significant performance gains, always exceeding those achieved by encoder scaling alone. We also identify label noise as a key challenge in STR, particularly in real-world data, which can limit the effectiveness of STR models. To address this, we propose Cloze Self-Distillation (CSD), a method that mitigates label noise by distilling a student model from context-aware soft predictions and pseudolabels generated by a teacher model. Additionally, we enhance the decoder architecture by introducing differential cross-attention for STR. Our methodology achieves state-of-the-art performance on 10 out of 11 benchmarks using only real data, while significantly reducing the parameter size and computational costs.
\end{abstract}    
\vspace{-0.9cm}
\section{Introduction} \label{sec:intro}

Scene Text Recognition (STR) aims to automatically transcribe text in natural scenes, enabling applications in autonomous driving \cite{zhang2020street}, augmented reality \cite{ouali2022text}, language translation \cite{vaidya2023show}, and assistive technologies. Unlike traditional Optical Character Recognition (OCR), which typically works with clean or scanned documents, STR faces unique challenges due to the diverse and uncontrolled nature of text in real-world environments. In particular, text in these settings can vary significantly in orientation, font style, shape, size, color, formatting, and aspect ratio. It often appears on complex backgrounds that may also include reflections, transparency, or occlusions. Furthermore, images might have poor quality, suffering from issues such as blurring, low resolution, and noise~\cite{long2021scene}.

\begin{figure}[t!]
\centering
\includegraphics[width=0.91\linewidth]{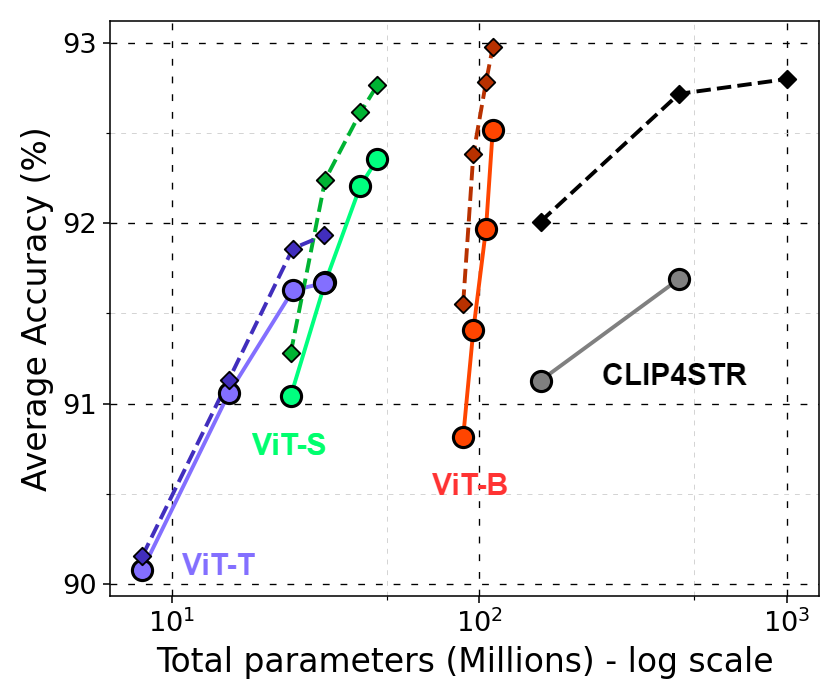}
\vspace{-1em}
\caption{\textbf{Average word accuracy} (\%) on $11$ STR benchmarks for the models with ViT-T, ViT-S and ViT-B vision encoders and $4$ different decoder sizes (see Sec. \ref{subsec:scaling_analysis}). Results are compared with the previous state-of-the-art model, CLIP4STR~\cite{zhao2023clip4str}. Results using \tb{Real} training dataset (3.3M images) are depicted with solid lines and circle markers, while results using \tb{RBU} training dataset (6.5M images) are shown with dashed lines and diamond markers. The x-axis represents the \tb{total number of model parameters} (in millions) on a logarithmic scale.}
\vspace{-1.2em}

\label{fig:plot_scaling}
\end{figure}




\noindent Recent research has led to notable performance improvements in STR by enhancing training methods~\cite{jiang2023revisiting}, deploying novel architectures \cite{zhao2023clip4str, atienza2021vision}, and exploring the effects of model scaling~\cite{rang2023large}. Despite these advancements, current STR models still face important challenges that limit their effectiveness. Our work is motivated by the following research question: 

\begin{quote}
\textit{What are the primary bottlenecks currently limiting STR, and what strategies can be employed to improve both accuracy and efficiency?}
\end{quote}

\noindent Throughout our analysis, we identify three important limitations: \tb{(i)} sub-optimal model scaling, \tb{(ii)} noisy labels in training data, and \tb{(iii)} architectural limitations within current model designs.

\noindent \tb{Sub-optimal model scaling.} Prior scaling analyses~\cite{rang2023large} have explored scaling laws for STR, demonstrating that increasing model size and data volume can lead to performance gains, following scaling trends similar to those observed in Natural Language Processing \cite{kaplan2020scaling}. In particular, the CLIP4STR~\cite{zhao2023clip4str} methodology, which leverages CLIP~\cite{radford2021learning} pre-training and integrates a cross-modal correction branch, achieves the best results at scale among all considered methods. However, \cite{rang2023large} also found that increasing the decoder depth in PARSeq~\cite{bautista2022scene} results in decreased performance, leading to an emphasis on encoder scaling. 

\noindent In this work, we provide an in-depth analysis of the effects of independently scaling the encoder and decoder components under different data volumes. Contrary to previous findings, we demonstrate that decoder scaling is indeed essential for achieving optimal STR performance. 

\noindent As illustrated in Fig. \ref{fig:plot_scaling}, increasing the decoder size provides substantial benefits for any visual encoder and results in more favorable scaling laws. Notably, proper model scaling alone is sufficient to surpass (on average) previous state-of-the-art performance without the need for CLIP pre-training or additional cross-modal branches. Furthermore, this approach substantially reduces the number of parameters and FLOPs.

\begin{figure}
\centering
\captionsetup{justification=centering}

\begin{subfigure}[b]{.25\linewidth}

 \caption{}\label{fig:c}
\includegraphics[width=\linewidth]{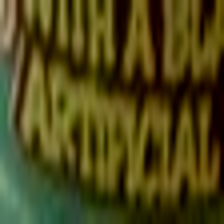}
\caption*{\scriptsize \tb{L}: \texttt{ARTIPICAL} \\ \tb{P}: \texttt{ARTIFICIAL}}
\end{subfigure}
\begin{subfigure}[b]{.25\linewidth}
 \caption{}\label{fig:c}
\includegraphics[width=\linewidth]{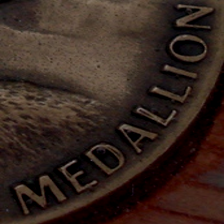}
\caption*{\scriptsize \tb{L}: \texttt{MEDALION} \\ \tb{P}: \texttt{MEDALLION}}
\end{subfigure}
\begin{subfigure}[b]{.25\linewidth}
 \caption{}\label{fig:c}
\includegraphics[width=\linewidth]{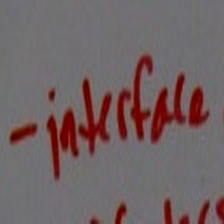}
\caption*{\scriptsize \tb{L}: \texttt{0} \\ \tb{P}: \texttt{-interface}}
\end{subfigure}

\begin{subfigure}[b]{.25\linewidth}
 \caption{}\label{fig:c}
\includegraphics[width=\linewidth]{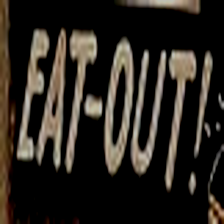}
\caption*{\scriptsize \tb{L}: \texttt{EAT-OUT} \\ \tb{P}: \texttt{EAT-OUT!}}
\end{subfigure}
\begin{subfigure}[b]{.25\linewidth}
 \caption{}\label{fig:c}
\includegraphics[width=\linewidth]{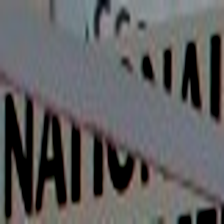}
\caption*{\scriptsize \tb{L}: \texttt{NA} \\ \tb{P}: \texttt{NATIONAL}}
\end{subfigure}
\begin{subfigure}[b]{.25\linewidth}
 \caption{}\label{fig:c}
\includegraphics[width=\linewidth]{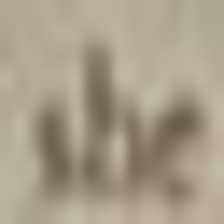}
\caption*{\scriptsize \tb{L}: \texttt{the} \\ \tb{P}: \texttt{she}}
\end{subfigure}

\captionsetup{justification=justified}
\vspace{-1em}
\caption{\tb{Examples of label inconsistencies and errors in the training set.} For each image, we show the ground truth label \tb{(L)} and the teacher-generated pseudolabel \tb{(P)}. Subfigures (a-c) illustrate typical label errors, such as spelling mistakes or missing characters. Subfigures (d,e) highlight label inconsistencies, where punctuation or occluded parts are not annotated. Subfigure (f) demonstrates a labelling error caused by severe degradation in the image quality.}
\vspace{-1.2em}
\label{fig:example_images_labels}
\end{figure}

\noindent \tb{Noisy labels in training data.} Our analysis indicates that, under some conditions, scaling the vision encoder may lead to diminishing returns or a decrease in accuracy, especially when STR models are trained on limited real data. In this context, we observe that text annotations of STR datasets often suffer from inconsistencies, errors and noise, which can negatively impact STR performance (Fig.~\ref{fig:example_images_labels}). To address this issue, we propose a novel \textit{Cloze Self-Distillation} (CSD) technique. In CSD, a model serving as a teacher, is first trained and used to generate predictions on training data. These predictions are then refined using a cloze-filling approach: each character is re-predicted using all other characters as textual context, resulting in more accurate, informative and context-aware soft predictions. We then distill the teacher into an identical student model on the same training set by employing the teacher's hard predictions as ground truth and a knowledge distillation term \cite{hinton2015distilling}  that minimizes the divergence between the student's limited-context predictions (obtained through permuted language modeling \cite{yang2019xlnet}) and the teacher's full-context cloze predictions. This technique enables the student to update its parameters with the richer, context-aware outputs of the teacher, while operating under the constrains of limited context. We provide empirical evidence demonstrating the effectiveness of CSD in mitigating label noise and inconsistencies, leading to substantial performance improvements.

\noindent \tb{Architectural limitations.} We extend our analysis on the decoder of our STR model by introducing additional  architectural improvements. Inspired by Differential Transformer \cite{ye2024differential}, we propose a novel Differential Permutation Language Decoder that employs Differential Cross-attention layers and SwiGLU activations~\cite{shazeer2020glu}, addressing the limitation of previous architecture in focusing on relevant context. 

\noindent Our contributions can be summarized as follows:
\begin{itemize}
    \item \tb{A detailed analysis of encoder-decoder scaling} for STR (Sec.~\ref{subsec:scaling_analysis}), demonstrating  substantial performance improvements with decoder-focused scaling, contrary to findings in previous studies.
    \item \tb{Cloze Self-Distillation (CSD)} technique that addresses label noise by leveraging context-rich cloze predictions (Sec. \ref{subsec:csd}), improving the robustness and performance of models across different data regimes. 
    \item \tb{An enhanced decoder architecture} that incorporates Differential Cross-Attention and SwiGLU activations (Sec. \ref{subsec:diff}), achieving further improvements in STR performance.
\end{itemize}

\noindent Through extensive empirical evaluation (Sec.~\ref{subsec:comparison_sota}), we demonstrate that the our enhanced decoder architecture and CSD, together with proper model scaling, consistently outperform previous approaches. Specifically, \tb{our STR method achieves the state-of-the-art performance on 10 out of 11 benchmarks}, with substantial reductions in parameter size and computational costs (FLOPs).

\section{Related work} \label{sec:related}
\textbf{Scene Text Recognition.} 
A branch of STR approaches relies on Connectionist Temporal Classification (CTC) \cite{graves2006connectionist}. These include approaches such as CRNN \cite{shi2016end}, DTRN \cite{he2016reading} and Star-Net \cite{liu2016star}, that utilize Convolutional Neural Networks (CNNs) and Recurrent Neural Networks (RNNs), as well as Rosetta \cite{borisyuk2018rosetta}. They utilize the character interactions using convolutions and recurrent structures. All these methods are trained with the CTC loss which enables to predict variable-length sequences without requiring explicit alignment. Another direction of approaches integrates attention mechanisms, as seen in RARE \cite{shi2016robust}, R2AM \cite{lee2016recursive}, ASTER \cite{shi2018aster} and DAN \cite{wang2020decoupled}, to capture the complex spatial dependencies of text characters. Similarly, VITSTR \cite{atienza2021vision} uses an encoder-only Vision Transformer \cite{dosovitskiy2020image} to encode the image patches that are directly classified into characters. A limitation of these approaches is that language modelling is not incorporated,
resulting in a weakness to strong perturbations and occlusions commonly encountered in STR.
To address this issue, a subsequent amount of methods incorporates context-aware mechanisms by integrating external or internal architectures,
such as NRTR \cite{sheng2019nrtr}, ABINet \cite{fang2021read}, TrOCR \cite{li2023trocr}, and PARSeq \cite{bautista2022scene}. In particular, PARSeq proposes to utilize an encoder-decoder transformer architecture and to train the model with an end-to-end scheme with permuted language modeling \cite{yang2019xlnet}. 
Similarly, DTrOCR \cite{fujitake2024dtrocr} uses a  decoder-only transformer (GPT-2 model \cite{radford2019language}) to directly decode image patches. Exploiting a pre-training on a large-scale simulated dataset and a fine-tuning step on real data, this method demonstrates state-of-the-art performance in many STR benchmarks.


\noindent \textbf{Empirical analyses.} Baek et al. \cite{baek2019wrong} examine the impact of training datasets on performance and inconsistencies in evaluation in the field of STR. Recently, Rang et al. \cite{rang2023large} investigate how model size, data volume, and computational resources affect the STR performance, revealing smooth power-law relationships between these factors and model accuracy. 

\noindent \textbf{Knowledge distillation (KD)}~\cite{hinton2015distilling} is a technique to enhance the model efficiency by replicating the knowledge of a complex \textit{teacher} model into a smaller \textit{student} model. In the context of STR, \cite{bhunia2021text} employs KD to unify STR and Handwriting Text Recognition models, while \cite{wang2023symmetrical} explore a symmetrical distillation strategy to capture the visual and linguistic knowledge of CLIP.

\section{Setup} \label{sec:setup}
\textbf{Notation.} We denote an input image as $\mb{x} \in \mathcal{X}$, where $\mathcal{X}$ is the image space, and a sequence of characters as  $\mb{y} = [y_1, y_2, \ldots, y_L] \in \mathcal{Y}$, where $\mathcal{Y}$ is the sequence space, $L$ is the sequence length and ($\forall i$) $y_i$ belongs to a fixed vocabulary $\mathcal{C}$ (character set). We use $\mb{y}_{<t} = [y_1, y_2, \ldots, y_{t-1}]$ to denote the subsequence of $\mb{y}$ previous to position $t$ and $\mb{y}_{\neq t} = [y_1, \ldots, y_{t-1}, y_{t+1}, \ldots, y_L]$ to denote the sequence $\mb{y}$ excluding the character in position $t$. In our mathematical formulation, we consider all sequences to be of the same length $L$. This can be achieved by right-padding shorter sequences with a special token \texttt{[PAD]}~$\in \mathcal{C}$. We use $\sigma(\cdot)$ to denote the softmax function and $D_{\text{KL}}(p || q)$ to denote the Kullback-Leiber divergence between distributions $p$ and $q$.

 
\noindent\textbf{Problem Formulation.} We formulate the STR problem as an image-conditioned generative language modeling task, where the objective is to model the conditional probability of a target characters sequence $\mb{y}$ given the input image $\mb{x}$. Let $\mathcal{D}$ represent the underlying data distribution over $\mathcal{X} \times \mathcal{Y}$ and let $p_{\theta}(\mb{y}|\mb{x})$ denote the probabilistic model, parametrized by $\theta$. The goal is to minimize the negative log-likelihood (NLL) of the sequence $\mb{y}$ given $\mb{x}$, formulated by:

\begin{equation}
    \min_{\theta} \mathbb{E}_{(\mb{x}, \mb{y}) \sim \mathcal{D}}[-\log p_{\theta}(\mb{y}|\mb{x})]
    \label{eq:maximum_likelihood} 
\end{equation}

\noindent
Since we do not have direct access to real data distribution $\mathcal{D}$, we approximate the objective using a finite dataset of i.i.d. samples $\{(\mb{x}_i, \mb{y}_i)\}_{i=1}^{n} \sim \mathcal{D}^n$, yielding:
\begin{equation}
    \min_{\theta}\frac{1}{n}\sum_{i=1}^n - \log p_\theta(\mb{y}_i|\mb{x}_i)
    \label{eq:maximum_likelihood_empirical} 
\end{equation}

\noindent
In standard language modeling, the model computes the full probability of the sequence $\mb{y}$ by conditioning each character $y_t$ on the previous sub-sequence $\mb{y}_{<t}$ that can be represented by:
\begin{equation}
    p_\theta(\mb{y}|\mb{x}) = \prod_{t=1}^{L} p_\theta(y_t|\mb{y}_{<t}, \mb{x})
    \label{eq:factorization}
\end{equation}

\noindent
This factorization enables the model to capture the left-to-right sequential dependencies of $\mb{y}$. During inference an output sequence $\hat{\mb{y}}$ can be predicted by iteratively selecting the most likely character $c$ over the character set $\mathcal{C}$. For each position $t \in [1,L]$, it can be formulated by:


\begin{equation}
\hat{y}_t = \arg\max_{c \in \mathcal{C}} p_\theta( y_t = c| \hat{\mb{y}}_{<t}, \mb{x})
\label{eq:greedy_decoding}
\end{equation}

\noindent
\textbf{Permutation Language Modeling (PLM)}. Initially introduced by~\cite{yang2019xlnet} to enable bidirectional context utilization in language models, PLM has been extended to the domain of STR by PARSeq~\cite{bautista2022scene}, to provide a flexible modeling approach. PLM generalizes the factorization of Eq.~\ref{eq:factorization} by considering multiple possible orders of character generation. Let $\Pi$ denote the set of all possible permutations of character indices $[1, 2, \ldots, L]$ for a sequence of length $L$. Then, the factorization based on the order induced by the permutation $\boldsymbol{\pi} \in \Pi$ is: 

\begin{equation}
    p_\theta(\mb{y}|\mb{x}) = \prod_{t=1}^{L} p_\theta(y_{\pi_t}|\mb{y}_{\boldsymbol{\pi}_{<t}}, \mb{x})
    \label{eq:factorization2}
\end{equation}

\noindent
By introducing this factorization in the NLL minimization problem of Eq. \ref{eq:maximum_likelihood}, the PLM objective becomes:

\begin{equation}
    \min_{\theta}
    \mathbb{E}_{\substack{(\mb{x}, \mb{y}) \sim \mathcal{D} \\ \boldsymbol{\pi} \sim \Pi}} \left[ \sum_{t=1}^L - \log p_\theta(y_{\pi_t}|\mb{y}_{\boldsymbol{\pi}_{<t}}, \mb{x}) \right]
    \label{eq:PLM_loglikelihood}
\end{equation}

\noindent
The empirical PLM counterpart of Eq.~\ref{eq:maximum_likelihood_empirical} can be easily derived using this equation. By removing the left-to-right constraint, PLM enables the model to use bidirectional information during training, enhancing its ability to handle diverse text layouts and ultimately improves the accuracy in STR~\cite{bautista2022scene}. In our model, we employ PLM in all experiments.

\noindent
\textbf{Cloze-filling refinement.} 
After PLM training, the model $p_\theta$  can be used to infer the sequences in any order, enabling the application of \textit{Cloze-filling refinement}. In this approach, first an initial prediction $\hat{\mb{y}}$ is made for an image $\mb{x}$, usually using the standard left-to-right decoding (Eq.~\ref{eq:greedy_decoding}). Later, each position $t$ in the sequence is re-predicted, given all the other characters $\hat{\mb{y}}_{\neq t}$ in order to obtain the the cloze-refined prediction $\hat{y}_t^{\text{cloze}}$ formulated by:

\begin{equation}
    \hat{y}_t^{\text{cloze}} = \arg\max_{c \in \mathcal{C}} p_\theta (y_t = c | \hat{\mb{y}}_{\neq t}, \mb{x})
\end{equation}


\section{Methodology} \label{sec:method}

In this work, we consider a STR model $p_\theta$ composed by an image encoder $E$ and a text decoder $D$. Given an input image $\mb{x}$, the encoder $E$ computes a sequence of vision tokens that constitute the latent representations $\mb{z} \in \mathcal{Z}$. Later, for a random permutation $\boldsymbol{\pi}$ and given $\mb{z}$, a sequence position $\pi_t$ and previous characters $\mb{y}_{\boldsymbol{\pi}_{<t}}$,  the decoder $D$ estimates the logits over the character set $\mathcal{C}$ in order to predict the $\pi_t^{\text{th}}$ character in the input image $\mb{x}$. In particular, our STR model can be formulated by:

\vspace{-0.4cm}
\begin{equation}
    p_\theta(y_{\pi_t} = c|\mb{y}_{\boldsymbol{\pi}_{<t}}, \mb{x}) = \sigma(D(E(\mb{x}),\pi_t,\mb{y}_{\boldsymbol{\pi}_{<t}}))_c
    \label{eq:model_enc_dec}
\end{equation}

\noindent where the subscript $c$ is used to indicate the index of the probability vector computed with the softmax function $\sigma(\cdot)$ associated to character $c$. Furthermore, the parameters $\theta = (\theta_E, \theta_D)$ denote both the encoder and decoder parameter sets, omitted from the Eq.~\ref{eq:model_enc_dec} for simplicity.

\begin{figure}[t]
\centering
\includegraphics[width=0.4\textwidth]{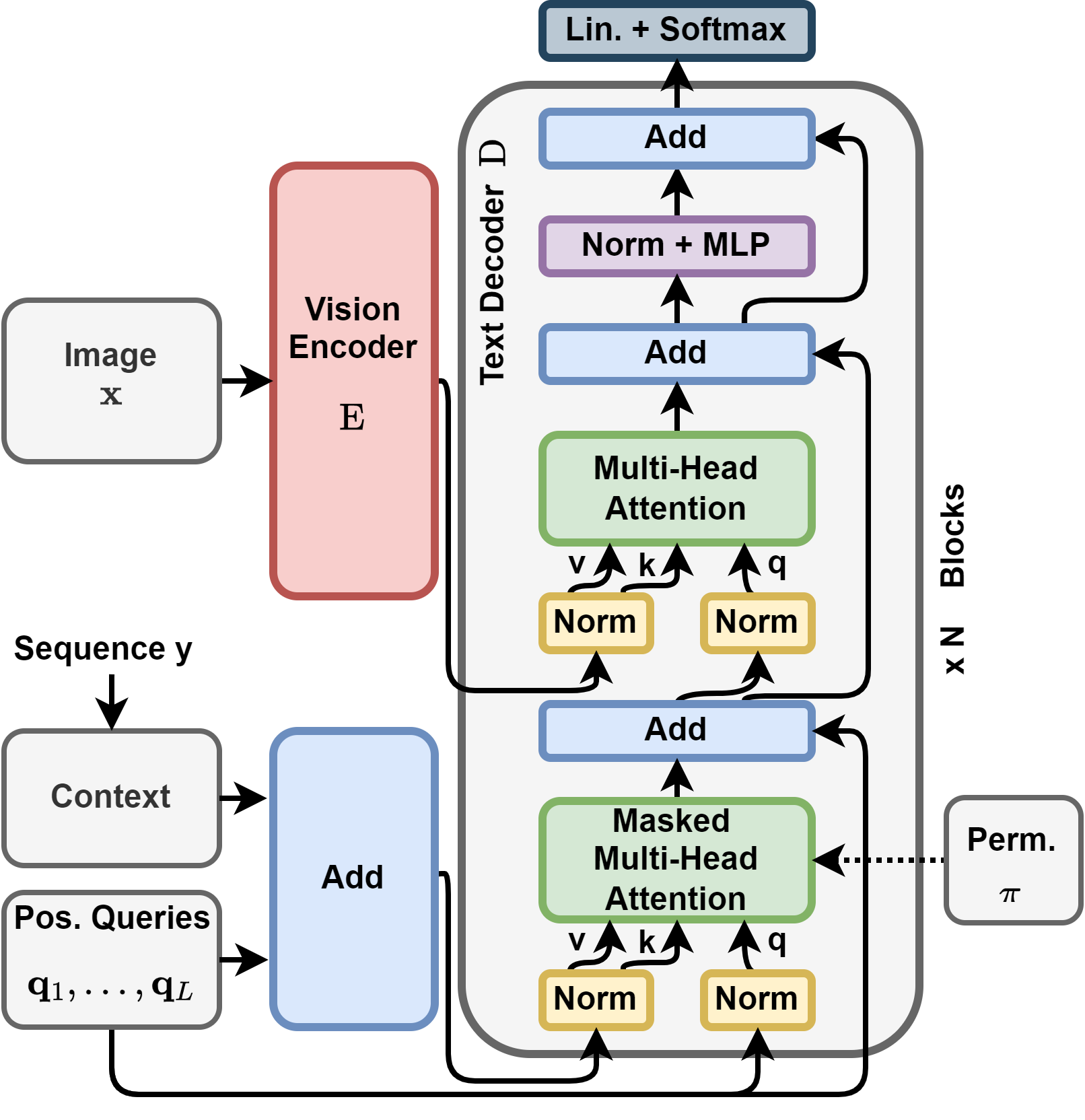}
\vspace{-1em}
\caption{\textbf{The overall architecture of our STR model}. Our model mainly consists of \textbf{Vision Encoder} $E$ and \textbf{Text Decoder} $D$. Details are given in the Sec.~\ref{sec:method}.}
\vspace{-1.2em}
\label{fig:architecture}
\end{figure}

\noindent
\textbf{Permutation Language Decoder (PLD)}. To enable PLM, the decoder $D$ is implemented with a specific transformer architecture that separates the query stream from the key-value stream, to account for its three inputs (Eq.~\ref{eq:model_enc_dec}) and to decode the sequence in any order. As shown in Fig.~\ref{fig:architecture}, each decoder block is composed by two Multi-Head Cross-Attention layers and by one MLP, with pre-normalization and skip connections. 
Sequence positions $[1, \ldots, L]$ are embedded into positional query vectors $[\mb{q}_{1}, \ldots, \mb{q}_{L}]$ that are the input of the query stream: to predict the character in position $\pi_t$, positional query $\mb{q}_{\pi_t}$ can be employed. Additionally, positional queries are also used as positional encoding and added to the context, i.e. the embedded sequence of previously predicted characters $\mb{y}_{\boldsymbol{\pi}_{<t}}$. Notably, this is introduced as input of the key-value stream in the first Cross-Attention layer, while vision tokens $\mb{z}$ are introduced in the second Cross-Attention.
Differently from previous approaches~\cite{bautista2022scene,rang2023large}, when using multiple blocks, we do not update the context or vision tokens due to observed performance degradation and increased computational complexity as we analyze in the supplementary material (Sec. \ref{sec:context_update}).
During training, all the positional queries and the complete ground truth context are utilized as inputs to enable parallelism, while the first Cross-Attention is masked to enforce the order of the input permutation $\boldsymbol{\pi}$ by generalizing the causal mask used in standard language modeling to any permuted order. For more details, refer to \cite{bautista2022scene}.







\subsection{Scaling Analysis}
\label{subsec:scaling_analysis}
We analyze how increasing model size can affect the final performance in our STR model, by specifically showing the individual effects of encoder and decoder scaling. To achieve this goal, we consider three image encoders with the same transformer architecture (Vision Transformer \cite{dosovitskiy2020image}) and pre-training scheme (on ImageNet21k \cite{deng2009imagenet}), but different parameter capacities: ViT-Tiny, ViT-Small, and ViT-Base. Additionally, we employ Permutation Language Decoders with four different sizes: PLD-Tiny, PLD-Small, PLD-Base, and PLD-Large. The details of these architectures in terms of hyperparameters, number of parameters and GFLOPs are presented in Tab.~\ref{tab:architecture_details}.
Additionally, we analyze the effects of training data volume to provide a comprehensive study and novel insights on how to scale STR models effectively in different data volumes.  

\setlength{\dashlinedash}{1pt}
\setlength{\dashlinegap}{2pt}

\begin{table}
  \centering
  
  \setcellgapes{-1pt}
  
  \makegapedcells
  \addtolength{\tabcolsep}{-0.3em}
  \begin{tabular}{clccccc}
    \toprule
    & & \tb{Blocks} & \tb{Dim} & \tb{Heads} & \tb{Params} & \tb{GFLOPs} \\
    \midrule
    \multirow{3}{*}{\rotatebox[origin=c]{90}{\scriptsize \tb{ENCODER}}} & ViT-Tiny & 12 & 192 & 3 & 5.5 M & 2.2 \\
    &  ViT-Small & 12 & 384 & 6 & 21.7 M & 8.6 \\
    &  ViT-Base & 12 & 768 & 12 & 85.8 M & 33.9 \\
   
    \midrule
    \multirow{5}{*}{\rotatebox[origin=c]{90}{\scriptsize \tb{DECODER}}} & PLD-Tiny & 1 & 384 & 6 & 2.5 M & 0.8 \\
    & PLD-Small & 1 & 768 & 12 & 9.6 M & 3.5 \\
    & PLD-Base & 2 & 768 & 12 & 19.1 M & 7.0 \\
    & PLD-Large & 3 & 768 & 12 & 28.8 M & 12.5 \\
    \cdashline{2-7}
    \noalign{\vskip 0.5ex}
    & PLD-Diff & 2 & 768 & 12 & 24.4 M & 7.1 \\
    \bottomrule
  \end{tabular} 
  \vspace{-0.8em}
  \caption{\textbf{Details of ViT encoders and PLD decoders} used in our scaling experiments. GFLOPs for the decoder refer to the average test sequence length $L=5.5$.}
  \vspace{-1.2em}
  \label{tab:architecture_details}
\end{table}

\subsection{Cloze Self-Distillation}
\label{subsec:csd}

Previous empirical evaluations \cite{bautista2022scene, rang2023large} have demonstrated that real datasets offer a better sample-efficiency for training STR models than synthetic datasets, as they are more closely aligned with the target distributions found in STR tasks.
However, despite their importance, real datasets often contain a large number of label errors and inconsistencies, which can adversely impact the performance of STR models, as qualitatively presented in Fig.~\ref{fig:example_images_labels}. We propose a novel technique, named \tb{Cloze Self-Distillation (CSD)}, to mitigate the impact of such errors and to improve the STR performance. In particular, CSD is motivated by two key observations:

\begin{itemize}
    \item After a complete training, the predictions of STR models are, in most cases, more accurate than the actual training labels (see Fig. \ref{fig:example_images_labels}).
    \item PLM allows to refine the predictions with the cloze-filling approach (end of Sec. \ref{sec:setup}) and to compute context-aware probabilities for each position $t$ in the sequence given all the other characters $\mb{\hat{y}_{\neq t}}$ as context.
\end{itemize}


 \noindent Given a dataset $\mathcal{S}_{\text{noise}}$ with potential label noise, CSD involves three main steps: \tb{(i)} a teacher STR model $p_{\theta_T}$ is fully trained on the noisy dataset $\mathcal{S}_{\text{noise}}$; \tb{(ii)} $p_{\theta_T}$ is employed to compute pseudolabels and context aware-logits with the cloze-filling refinement for the dataset $\mathcal{S}_{\text{noise}}$; \tb{(iii)} a new student model $p_{\theta_S}$ (with the same architecture and size of the initial model) is distilled from the teacher.
 Hence, teacher pseudolabels are used instead of the ground truth annotations to minimize the negative log likelihood (NLL) objective of Eq. \ref{eq:PLM_loglikelihood} and an additional Knowledge Distillation (KD) loss term is introduced to minimize the divergence between the context-aware soft predictions of the teacher (obtained with cloze-filling) and the partial-context predictions of the student (obtained with PLM), as it is illustrated in Fig.~\ref{fig:csd_flow}. 
 Formally, the KD term can be formulated by:
 \vspace{-1.5em}

  \begin{figure}[t]
\centering
\includegraphics[width=0.93\linewidth]{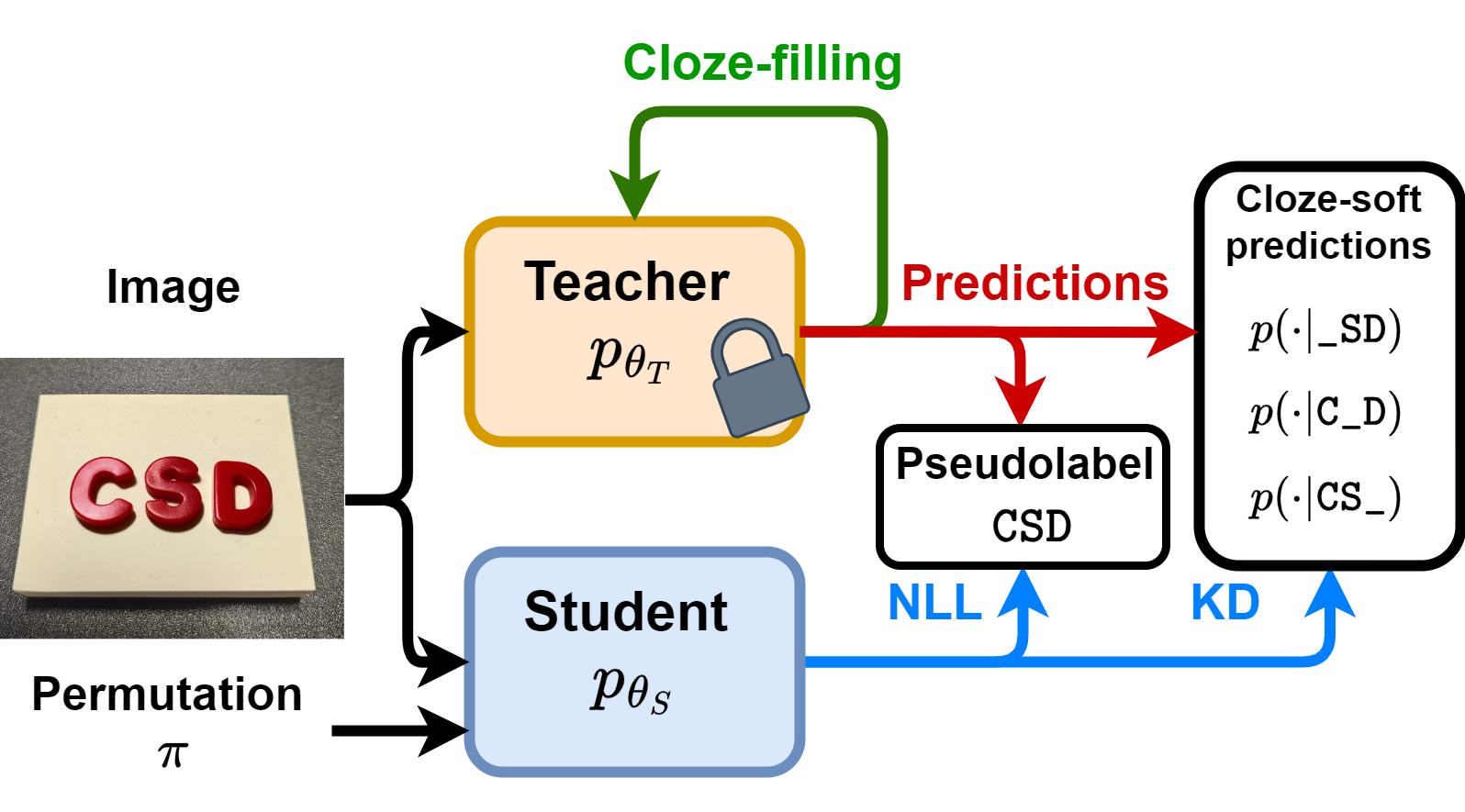}
\vspace{-1em}
\caption{\tb{Flow of Cloze Self-Distillation (CSD).} Pseudolabels and soft predictions of a fixed teacher model, obtained with the cloze-filling approach, are distilled into a student model by minimizing the negative log likelihood (NLL) and the knowledge distillation (KD) objective, presented in Eq. \ref{eq:CSD_objective}.}
\vspace{-1.2em}
\label{fig:csd_flow}
\end{figure}
  

 \begin{align}
     \text{KD}_{\boldsymbol{\pi}, t}(\mb{x}, \mb{y}) &=  D_{\text{KL}}\left( p^{\tau}_{\theta_T}(\cdot|\mb{y}_{\boldsymbol{\pi}_{\neq t}}\ \mb{x}) \middle|\middle| p^{\tau}_{\theta_S}(\cdot | \mb{y}_{\boldsymbol{\pi}_{<t}}, \mb{x})\right) 
     \label{eq:knowledge_distillation}
 \end{align}
\noindent
where the superscript $\tau$ is used to indicate that the logits of the models are scaled with temperature $\tau$ before computing the softmax outputs.
We remark that the teacher soft-predictions are computed given the full context, $\mb{y}_{\boldsymbol{\pi}_{\neq t}}$, while the student outputs are computed with the standard context of PLM, $\mb{y}_{\boldsymbol{\pi}_{<t}}$. This intuitively makes the task more challenging for the student, allowing it to effectively distill the knowledge from the teacher and to outperform its performance. The overall CSD objective is:

\begin{equation}
    \min_{\theta}  \mathbb{E}_{\substack{(\mb{x}, \mb{y}) \sim \mathcal{D} \\ \boldsymbol{\pi} \sim \Pi \\ t \sim [1, L]}} \left[ - \log p_\theta(y_{\pi_t}|\mb{y}_{\boldsymbol{\pi}_{<t}}, \mb{x}) + \alpha\text{KD}_{\boldsymbol{\pi}, t}(\mb{x}, \mb{y})  \right] 
    \label{eq:CSD_objective}
\end{equation}

\noindent
where the expectation $\mathbb{E}[.]$ is with respect to the data samples $(\mb{x}, \mb{y})$, the permutations $\boldsymbol{\pi}$ and the position $t$ in the sequence. Furthermore, $\alpha$ is the hyperparameter to define the contribution of distillation term to the objective function.

\subsection{Differential Decoder}
\label{subsec:diff}

\begin{figure}[t]
\centering
\includegraphics[width=0.4\textwidth]{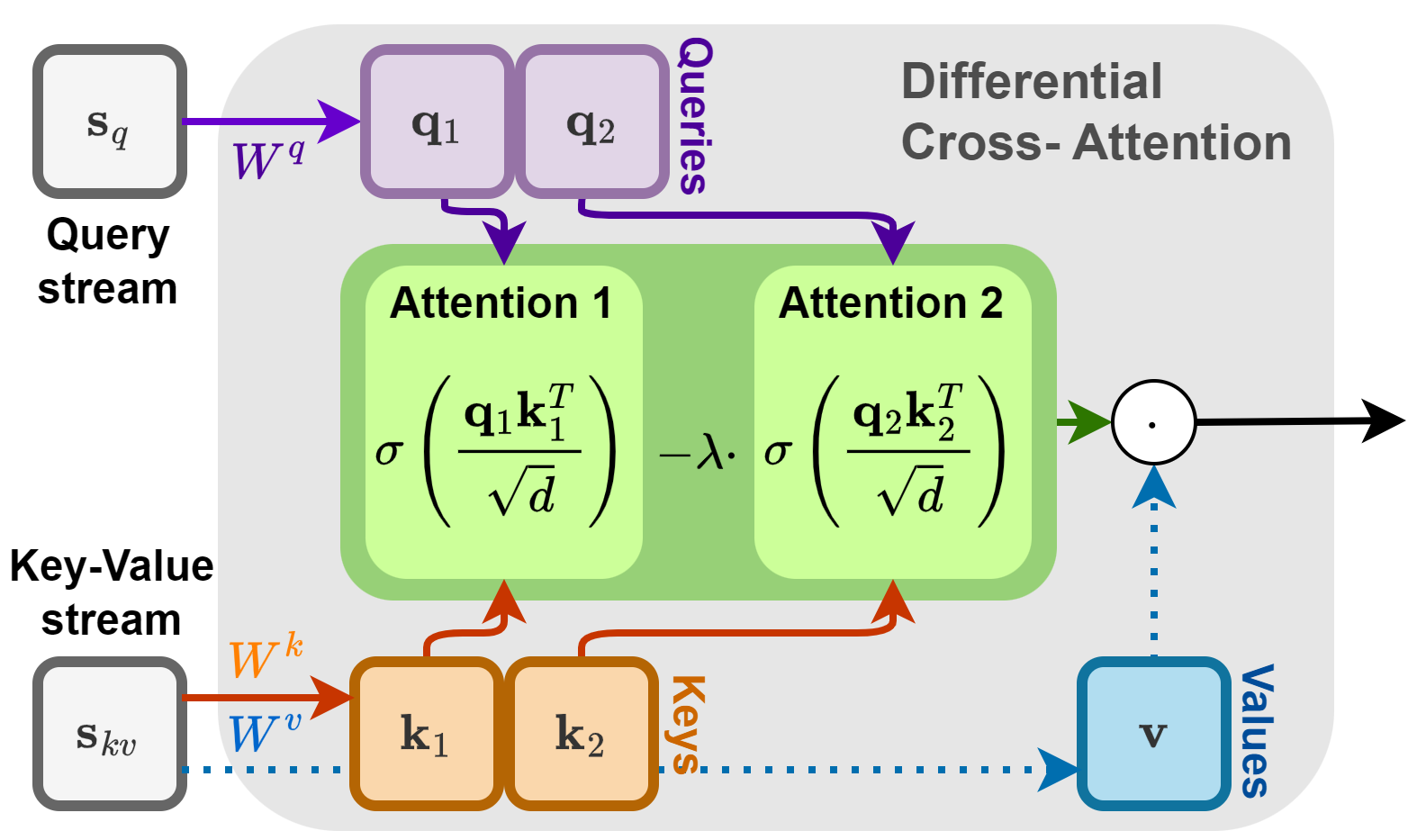}
\vspace{-0.5em}
\caption{\textbf{Differential Cross-Attention}  used in our PLD decoder. For simplicity, the diagram shows a single head.}
\vspace{-2em}
\label{fig:diff_architecture}
\end{figure}

For the decoder $D$, we introduce a Differential Cross-Attention mechanism, inspired from the Differential Self-Attention proposed by \cite{ye2024differential} for NLP. Intuitively, Differential Cross-Attention computes two separate softmax cross-attention maps and subtracts them to cancel out common-noise mode~\cite{ye2024differential}. This enables the Cross-Attention mechanism to focus more on relevant context and vision tokens than noisy-representations (visual examples are presented in the supplementary material, Sec.~\ref{sec:pretraining}).
\noindent Formally, let $d$ be the inner dimension of the transformer, $h$ be the number of heads, such that $d_h = \frac{d}{h} \in \mathbb{N}$ is the dimension of each head. Given an input sequence of $L_q$ tokens for the query stream $\mb{s}_q$ and an input sequence of $L_{kv}$ tokens for the key-value stream $\mb{s}_{kv}$, for the attention head $i \in [1, \ldots, h]$, the differential cross-attention operation is formulated by:

 \begin{align}
        \text{Cross-Att}_{\text{DIFF}}^{i}(\mb{s}_q, \mb{s}_{kv}) = \left( \sigma\left( \frac{\mb{q}_1 \mb{k}_1^T}{\sqrt{d}} \right) - \lambda \sigma\left( \frac{\mb{q}_2 \mb{k}_2^T}{\sqrt{d}}  \right) \right) \mb{v}
     \label{eq:ca_diff}
 \end{align}

\noindent
where the queries, keys and values in the attention operation can be represented as $\mb{q} = [\mb{q}_1, \mb{q}_2] = \mb{s}_q W^q_i$, $\mb{k} = [\mb{k}_1, \mb{k}_2] = \mb{s}_{kv} W^k_i$, and $\mb{v} = \mb{s}_{kv} W^v_i$, where $i$ indicates the head index and $W^q_i$, $W^k_i$, $W^v_i$ are the projection matrices and $[\cdot, \cdot]$ indicates a concatenation operation. $\lambda$ is a scalar parameter shared across heads of the same layer, that is re-parametrized by following~\cite{ye2024differential}. The outputs of the heads are separately normalized using RMSNorm, concatenated and multiplied with an output projection matrix $W^o$. In our Differential Decoder, we replace all traditional Multi-Head Cross-Attention layers with Multi-Head Differential Cross-Attention layers and utilize MLP with SwiGLU~\cite{shazeer2020glu}. Differently from \cite{ye2024differential} RoPE \cite{su2024roformer} is not applied.
\noindent A diagram of the Differential Cross-Attention is presented in Fig.~\ref{fig:diff_architecture}.

\vspace{-0.2cm}
\section{Results} \label{sec:results}
\vspace{-0.1cm}
\subsection{Datasets}

In this work, for training, we use a set of Real datasets commonly used in the literature and a superset RBU. For evaluation, we employ $11$ test benchmark datasets. 

\noindent \textbf{Real Dataset} ($3.3$M images). It is a large scale collection of real datasets including COCO-Text~\cite{veit2016coco}, RCTW17~\cite{shi2017icdar2017}, Uber-Text~\cite{zhang2017uber}, ArT~\cite{chng2019icdar2019}, LSVT~\cite{sun2019icdar}, MLT19~\cite{nayef2019icdar2019}, TextOCR~\cite{singh2021textocr}, ReCTR~\cite{zhang2019icdar} and OpenVINO~\cite{krylov2021open}. These datasets have samples that cover challenging cases for low-resolution, occluded, curved and rotated text. The detail analysis of these datasets is presented in~\cite{jiang2023revisiting}. 

\noindent \textbf{RBU} ($6.5$M images). It is the combination of Real dataset (R), the training split of benchmark datasets (B) and a subset of Union14M-L (U). The Benchmark split includes IIIT5K~\cite{mishra2012scene}, Street View Text (SVT)~\cite{wang2011end}, ICDAR13~\cite{karatzas2013icdar} and ICDAR15~\cite{karatzas2015icdar}, while the U split includes a subset of approximately $3.5$M images from Union14-L~\cite{jiang2023revisiting}.

\noindent \textbf{Test Benchmarks}. We evaluate STR models on the $6$ most widely used benchmarks in the literature:  ICDAR13 (IC13)~\cite{karatzas2013icdar}, IIIT5K~\cite{mishra2012scene}, and Street View Text (SVT)~\cite{wang2011end} for \textit{Regular Text} recognition; and CUTE80 (C80)~\cite{risnumawan2014robust}, ICDAR15 (IC15)~\cite{karatzas2015icdar} and Street View Text-Perspective (SVT-P)~\cite{phan2013recognizing} for \textit{Irregular Text} recognition. The IC13 benchmark includes subsets of $857$ images (IC13-857) and a subset of $1\,015$ images (IC13-1015), while IC15 include subsets of $1\,811$ images (IC15-1811) and $2\,077$ images (IC15-2077). Additionally, we report performance on two datasets designed to evaluate robustness for occlusions: Heavily Occluded Scene Text (HOST) and Weakly Occluded Scene Text (WOST)~\cite{wang2021two}. To further ensure comprehensive evaluation, we also include more recent and larger benchmarks, COCO-Text~\cite{veit2016coco}, ArT~\cite{chng2019icdar2019} and Uber-Text~\cite{zhang2017uber}. 

\subsection{Experiment Settings}
\textbf{Pre-processing.} For training, input images are augmented using RandAugment~\cite{cubuk2020randaugment} using $3$ different layers with magnitude $5$. By following \cite{bautista2022scene}, \texttt{Sharpeness} augmentation is excluded, while \texttt{GaussianBlur} and \texttt{PoissonNoise} are added. Images are then resized to $224 \times 224$ and their pixel values are normalized to the interval $[-1, 1]$. Since our vision encoder $E$ is based on ViT, images are converted into patches of size $16 \times 16$ pixels. Following previous works and to provide comparable results, we set the maximum sequence length $L$ to $25$ and we consider a set $\mathcal{C}$ of 94 characters during training (mixed-case alphanumeric and punctuation marks) and a set of 36 characters (lowercase alphanumeric) during test.

\noindent \textbf{Training protocol.} Model parameters are optimized with a global batch size of 1024 using AdamW \cite{loshchilov2017decoupled} with $\beta_1 = 0.9$, $\beta_2 = 0.95$, $0.1$ weight decay and gradients are clipped to $1.0$. The learning rate follows a One-Cycle \cite{smith2019super} schedule with a maximum value of $0.01$ and $3\,300$ warm-up steps, both for encoder and decoder parameters. We train each STR model for $110$K steps, corresponding to approximately $35$ epochs for Real (R) dataset, $17.5$ for RBU. To train the model with Permutation Language Modeling, we follow \cite{bautista2022scene} and employ the causal and anti-causal permutations (left-to-right and right-to-left), together with 4 random permutations sampled differently for each batch. For CSD, the same training schedule is employed both for the teacher and the student and we use $\alpha = 0.1$ and $\tau = 2.0$. 

\noindent \textbf{Evaluation metrics.} Following previous works, we evaluate our STR model using the \textit{word accuracy}, where a predicted sequence is considered correct only if all characters match the ones of the ground truth label. To make a comparison with the results reported by previous works~\cite{zhao2023clip4str, rang2023large}, we provide two separate aggregate scores: the average accuracy across all $11$ benchmarks, considering the subsets IC13-1015 and IC15-1811 (it is referred as \tb{AVG}$_{11}$) and the weighted average of the 6 common benchmarks with the subsets IC13-857 and IC15-1811 (it is referred as \tb{wAVG}$_{6}$).

\subsection{Encoder-Decoder Scaling results}
\label{subsec:scaling_results}
In this section, we analyze the impact of scaling both the vision encoder and the text decoder in our STR model. In Tab. \ref{tab:model_scaling}, we present the average word accuracy \tb{AVG}$_{11}$ achieved with various encoder and decoder configurations (introduced in Sec.~\ref{subsec:scaling_analysis}). Furthermore, Fig.~\ref{fig:plot_scaling} provides a visual comparison of these results, highlighting how proper model scaling can outperform the previous state-of-the-art (CLIP4STR \cite{zhao2023clip4str}) using significantly less parameters.

\noindent \textbf{Encoder scaling.} From the results, scaling the vision encoder from ViT-T to ViT-S significantly boosts the accuracy across both Real and RBU datasets, and for all decoder configurations. However, further scaling from ViT-S to ViT-B, shows a different effect: when data is abundant (on RBU), the larger encoder improves performance with all decoders, but on the Real dataset with smaller decoders, ViT-B decreses the performance compared to ViT-S. Part of this behavior can be explained due to the label noise sensitivity of ViT-B (when paired with a small decoder). In Subsection~\ref{subsec:csd_results}, we will show that the impact of label noise can be mitigated by our CSD technique.

\begin{table}
  \centering
  \resizebox{0.9\linewidth}{!}{
  \begin{tabular}{@{}lccccc@{}}
    \toprule
     & \textbf{Dataset} & \tb{PLD-T} & \tb{PLD-S} & \tb{PLD-B} & \tb{PLD-L} \\
    \cmidrule(lr){2-2} \cmidrule(lr){3-6}
    \multirow{2}{*}{\textbf{ViT-T}} & \tb{\small{Real}} & 90.08 & 91.06  & 91.63 & 91.67 \\
                                    & \dgt{\tb{\small{RBU}}}  & \dgt{90.15} & \dgt{91.13}  & \dgt{91.86} & \dgt{91.93} \\ 
    \cmidrule(lr){2-2} \cmidrule(lr){3-6}
    
    \multirow{2}{*}{\textbf{ViT-S}} & \tb{\small{Real}} & 91.04 & 91.67 & 92.21 & 92.36 \\
                                    & \dgt{\tb{\small{RBU}}}  & \dgt{91.28} & \dgt{92.24} & \dgt{92.62} & \dgt{92.77} \\ 
    
    \cmidrule(lr){2-2} \cmidrule(lr){3-6}
    \multirow{2}{*}{\textbf{ViT-B}} & \tb{\small{Real}} & 90.81 & 91.41 & 91.97 & 92.52 \\
                                    & \dgt{\tb{\small{RBU}}}  & \dgt{91.55} & \dgt{92.38} & \dgt{92.78} & \dgt{92.98} \\
    
    \bottomrule
  \end{tabular}}
  \vspace{-0.6em}
  \caption{\tb{Encoder-Decoder Scaling.} Average word accuracy (\%) on the 11 benchmarks (\tb{AVG}$_{11}$) for different encoder-decoder configurations trained on Real or RBU dataset.}
  \label{tab:model_scaling}
  \vspace{-0.6em}
\end{table}


\begin{table}
 \centering
 
 \resizebox{0.99\linewidth}{!}{
 \addtolength{\tabcolsep}{-0.3em}
 \begin{tabular}{lccccccccc}
    \toprule
    & \multicolumn{3}{c}{\tb{ViT-T}} & \multicolumn{3}{c}{\tb{ViT-S}} & \multicolumn{3}{c}{\tb{ViT-B}} \\
     \cmidrule(lr){2-4} \cmidrule(lr){5-7}  \cmidrule(lr){8-10} 
    \tb{P} & \xmark & \cmark & \cmark & \xmark & \cmark & \cmark & \xmark & \cmark & \cmark \\
    \tb{KD} & \xmark & \xmark & \cmark & \xmark & \xmark & \cmark & \xmark & \xmark & \cmark \\
    \cmidrule(lr){2-4} \cmidrule(lr){5-7}  \cmidrule(lr){8-10} 
    \tb{AVG}$_{11}$ & 91.6 & \tu{91.8} & \tb{91.9} & 92.2 & \tu{92.4} & \tb{92.5} & 92.0 & \tu{92.3} & \tb{92.5} \\
    \bottomrule
 \end{tabular}}
 
 \vspace{-0.6em}
 \caption{\tb{Effects of pseudolabels and KD.} Average word accuracy (\%) on 11 benchmarks (\tb{AVG}$_{11}$) using the Real dataset with standard supervised training, pseudolabels (\tb{P}) and Knowledge Distillation (\tb{KD}) with the cloze soft probabilities. Results are shown for different encoders paired with the base decoder (PLD-B).}
 \label{tab:csd_results_pkd}
 \vspace{-0.6em}
\end{table}

\begin{table}
  \centering
  \addtolength{\tabcolsep}{-0.0em}
  \begin{tabular}{lccccc}
  \toprule
  & \multicolumn{4}{c}{\tb{Real}} & \tb{RBU} \\
  & \textbf{10\%} & \textbf{25\%} & \textbf{50\%} & \textbf{100\%} & \textbf{200\%} \\
  \cmidrule(lr){2-5} \cmidrule(lr){6-6} 
  \textbf{Sup.} & 86.9 & 89.9 & 90.8 & 92.0 & 92.8 \\
  \textbf{CSD} & 89.1 & 91.1 & 91.7 & 92.5 & 93.2 \\
  \bottomrule
  \end{tabular}
  \vspace{-0.6em}
  \caption{\tb{Benefits of CSD.} Average word accuracy (\%) on the 11 benchmarks (\tb{AVG}$_{11}$) of ViT-B and PLD-B scaling the data samples from 0.33M (10\%) to 6.5M (200\%). Standard supervised training (\tb{Sup.}) is compared to our approach (\tb{CSD}). }
  \label{tab:data_scaling}
  \vspace{-0.6em}
\end{table}

\begin{table}
  \centering
  
  \resizebox{0.99\linewidth}{!}{
  \addtolength{\tabcolsep}{-0.3em}
  \begin{tabular}{lcccccc}
    \toprule
    \multirow{2}{*}{\textbf{Decoder}} &  \multirow{2}{*}{\tb{Params}} & \multirow{2}{*}{\textbf{GFLOPs}} & \multicolumn{2}{c}{\tb{Real}} & \multicolumn{2}{c}{\tb{RBU}} \\
    & & & \tb{AVG}$_{11}$ &\tb{wAVG}$_{6}$ & \tb{AVG}$_{11}$& \tb{wAVG}$_{6}$ \\
    \cmidrule(lr){4-5}\cmidrule(lr){6-7}
    PLD-B & 104.9 M & 40.9 & 92.5 & 97.3 & 93.2 & 97.5\\
    PLD-D & 110.2 M & 41.0 & \tb{92.7} & \tb{97.4} & \tb{93.3} & \tb{97.6} \\

    \bottomrule
  \end{tabular}}
  \vspace{-0.6em}
  \caption{\tb{Benefits of Differential Decoder (PLD-D).} \tb{AVG}$_{11}$ and \tb{wAVG}$_{6}$ of ViT-B paired with the standard base decoder (PLD-B) and the differential decoder (PLD-D), trained on Real or RBU dataset. Parameters and GFLOPs refer to the full encoder-decoder architecture, considering the average test sequence length of 5.5.}
  \label{tab:model_diff}
  \vspace{-1.5em}
\end{table}

\begin{table*}
  \centering
  \resizebox{0.99\textwidth}{!}{
  \addtolength{\tabcolsep}{-0.1em}
  \begin{tabular}{lllcccccccccccccc}
    \toprule
    & & & \multicolumn{4}{c}{\textbf{Regular text}} & \multicolumn{4}{c}{\textbf{Irregular Text}} & \multicolumn{2}{c}{\textbf{Occluded Text}} & \multicolumn{3}{c}{\textbf{Other}} & \\ 
     \cmidrule(lr){4-7} \cmidrule(lr){8-11} \cmidrule(lr){12-13} \cmidrule(lr){14-16}
    \multirow{2}{*}{\textbf{Method}} & \multirow{2}{*}{\textbf{Data}} & \multirow{2}{*}{\textbf{Params}} & \multicolumn{2}{c}{\textbf{IC13}} & \textbf{IIIT5k} & \textbf{SVT} & \textbf{C80} & \multicolumn{2}{c}{\textbf{IC15}} & \textbf{SVTP} & \textbf{HOST} & \textbf{WOST} & \textbf{ArT} & \textbf{COCO} & \textbf{Uber} & \textbf{AVG}$_{11}$\\
    
    & & & 857 & 1015 & 3000 & 647 & 288 & 1811 & 2077 & 645 & 2416 & 2416 & 34k & 9825 & 89.5k & \\
    \cmidrule(lr){1-3} \cmidrule(lr){4-7} \cmidrule(lr){8-11} \cmidrule(lr){12-13} \cmidrule(lr){14-16} \cmidrule(lr){17-17}

    VITSTR-S \cite{atienza2021vision} & Real &  21.7 M & 97.6       & 97.7      & 98.1      & 95.8      & 96.1      & 88.4      & 87.1      & 91.4      & 64.5\sstar     & 77.9\sstar     & 81.1      & 74.1      & 78.2  & 85.8\\
    CRNN \cite{shi2016end}            & Real &  8.5 M &  94.1      & 94.5      & 94.6      & 90.7      & 89.1      & 82.0      & 78.5      & 80.6      & -         & -         & 66.8      & 62.2      & 51.0  & - \\
    TRBA \cite{baek2019wrong}         & Real &  49.6 M &  97.6      & 97.6      & 98.6      & 97.0      & 97.7      & 89.8      & 88.7      & 93.7      & -         & -         & 82.5      & 77.5      & 81.2  & -\\
    ABINET \cite{fang2021read}        & Real &  23.5 M &  98.0      & 97.8      & 98.6      & 97.8      & 97.7      & 90.2      & 88.5      & 93.9      & 72.2\sstar     & 85.0\sstar     & 81.2      & 76.4      & 71.5  & 87.5\\
    PARSeq \cite{bautista2022scene}  & Real &  22.5 M &  98.3      & 98.4      & 99.1      & 97.9      & 98.3      & 90.7      & 89.6      & 95.7      & 74.4\sstar     & 85.4\sstar     & 84.5      & 79.8      & 84.5 & 89.9\\
    CLIP4STR-B \cite{zhao2023clip4str} & Real &  158 M &  98.4\sdag     & 98.3      & 99.2      & 98.3      & \tb{99.3} & 91.4      & 90.6      & 97.2      & 77.5      & 87.5      & 85.8      & 81.1      & 86.8 & 91.1\\
    CLIP4STR-L \cite{zhao2023clip4str} & Real &  446 M &  98.5\sdag     & \tu{98.5} & \tb{99.5} & \tu{98.5} & \tu{99.0} & 91.3      & 90.8      & 97.4      & 79.8      & 89.2      & 85.9      & 81.9      & 87.6 & 91.7\\
    
    \cmidrule(lr){1-3} \cmidrule(lr){4-7} \cmidrule(lr){8-11} \cmidrule(lr){12-13} \cmidrule(lr){14-16} \cmidrule(lr){17-17}
    
    \tb{CSD-S (ours)} & Real &  40.8 M &  \tu{99.1} & \tb{98.8} & \tu{99.4} & \tu{98.5} & \tu{99.0} & 91.9      & 91.3      & 97.5      & \tu{83.5} & \tb{90.9} & \tb{86.2} & \tu{82.7} & 89.6 & \tu{92.5}\\
    \tb{CSD-B (ours)} & Real &  104.9 M &  \tb{99.2} & \tb{98.8} & \tu{99.4} & 98.0      & \tu{99.0} & \tb{92.5} & \tu{91.6} & \tu{97.8} & \tb{83.6} & 90.0      & \tb{86.2} & \tb{82.8} & \tu{89.7} & \tu{92.5} \\
    \tb{CSD-D (ours)} & Real &  110.2 M &  99.0      & \tb{98.8} & 99.3      & \tb{99.1} & \tb{99.3} & \tu{92.4} & \tb{91.7} & \tb{98.1} & \tb{83.6} & \tu{90.8} & \tu{86.1} & 82.6      & \tb{89.8} & \tb{92.7}\\
    
    \cmidrule[1pt](lr){1-3} \cmidrule[1pt](lr){4-7} \cmidrule[1pt](lr){8-11} \cmidrule[1pt](lr){12-13} \cmidrule[1pt](lr){14-16} \cmidrule[1pt](lr){17-17}
    
    CLIP4STR-B \cite{zhao2023clip4str} & RBU  &  158 M &  -         & 98.6      & \tu{99.5} & 98.3      & 99.0      & 91.4      & 91.1      & 98.0      & 79.3      & 88.8      & 85.8      & 81.3      & 92.1 & 92.0\\
    CLIP4STR-L \cite{zhao2023clip4str}& RBU  &  446 M &  -         & \tu{99.0} & \tb{99.6} & 98.6      & \tb{99.7} & 91.9      & 91.4      & \tu{98.1} & 81.1      & 90.6      & \tb{86.4} & 82.7      & 92.2 & 92.7\\
    CLIP4STR-H \cite{zhao2023clip4str}& RBU  &  1 B &  -         & 98.9      & \tu{99.5} & \tu{99.1} & 99.0      & 91.7      & 91.0      & 98.0      & 82.6      & \tu{90.9} & \tb{86.4} & 83.0      & 91.7 & 92.8\\
    
    \cmidrule(lr){1-3} \cmidrule(lr){4-7} \cmidrule(lr){8-11} \cmidrule(lr){12-13} \cmidrule(lr){14-16} \cmidrule(lr){17-17}
    
    \tb{CSD-S (ours)} & RBU  &  40.8 M  &  98.7      & 98.6      & 99.2      & 98.8     & 99.0      & 92.2      & 91.7      & 97.8      & \tb{84.3} & 89.5      & \tu{86.3}  & 82.9     & 91.7 & 92.8\\
    \tb{CSD-B (ours)} & RBU  &  104.9 M &  \tu{98.8} & 98.7      & \tu{99.5} & 98.8     & 99.3      & \tu{92.6} & \tb{92.2} & \tb{98.3} & 84.2      & \tb{91.2} & \tb{86.4} & \tb{83.4} & \tu{93.1} & \tu{93.2}\\
    \tb{CSD-D (ours)} & RBU  &  110.2 M &  \tb{99.2} & \tb{99.2} & \tu{99.5} & \tb{99.2} & \tb{99.7} & \tb{92.7} & \tu{91.9}  & \tu{98.1} & \tb{84.3} &  90.6     & \tb{86.4} & \tu{83.1} & \tb{93.2} & \tb{93.3}\\
    
    
    \bottomrule
  \end{tabular}}
  \vspace{-0.8em}
  \caption{\tb{Comparison with state-of-the-art methods}. The word accuracy (\%) of our models trained with CSD is compared with state-of-the-art approaches both for the Real and RBU training datasets. Results marked with \sstar are from \cite{zhao2023clip4str}, results marked with \sdag are from \cite{rang2023large}. The best results are highlighted in \tb{bold}, while second-best results are \tu{underlined}.}
  \label{tab:sota}
  \vspace{-1.2em}
\end{table*}

\noindent \textbf{Permutation Language Decoder scaling.} Our results demonstrate that scaling the decoder is more parameter (Fig.~\ref{fig:plot_scaling}) and computational (Sec.~\ref{sec:gflops}) efficient than scaling the encoder only, leading to more favorable scaling laws than previous state-of-the-art approaches. Using the larger RBU dataset as a reference, increasing the encoder from ViT-T to ViT-B yields an average improvement (across decoders) of $1.16\%$ \tb{AVG}$_{11}$ with an additional $80.3$M parameters. In contrast, scaling the decoder from PLD-T to PLD-L results in an average improvement (across encoders) of $1.56\%$ \tb{AVG}$_{11}$ with only $26.3$M parameter increase. Furthermore, on RBU, ViT-B paired with PLD-L ($114.6$M total parameters) obtains an average accuracy \tb{AVG}$_{11}$ of $92.98\%$ surpassing the $92.80\%$ accuracy of CLIP4STR-H (1B parameters).
\noindent Similar trends can be observed also on the Real dataset, where ViT-T, ViT-S and ViT-B configurations achieve notable performance gains when the decoder size is increases, and, the transition from PLD-T to PLD-L provides $+1.59\%$, $+1.32\%$ and $+1.71\%$, respectively.



\subsection{Cloze Self-Distillation results}
\label{subsec:csd_results}
Table~\ref{tab:csd_results_pkd} presents the average word accuracy (\tb{AVG}$_{11}$) obtained with different training procedures: standard supervised training, training on teacher pseudolabels (P) and CSD (pseudolabels and Knowledge Distillation (KD)). In this experiment, we utilize ViT of varying sizes (Tiny, Small, Base) as encoders, paired with the base-size decoder (PLD-B). Notably, incorporating teacher pseudolabels during training significantly enhances the performance, since it reduces the label errors and inconsistencies in real datasets. Moreover, integrating the Knowledge Distillation component based on context-aware probabilities (computed with the cloze-filling approach) further strengthens the regularization effects, resulting in an additional performance gain. The superiority of CSD is evident also in Table~\ref{tab:data_scaling}, which reports the average accuracy of ViT-Base with PLD-B when scaling the data from $10\%$ to $100\%$ of the Real dataset, as well as on RBU (which represents approximately $200\%$ of the Real dataset). Compared to the baseline method of standard supervised training, CSD consistently provides notable improvements at all data scales. Our technique achieves $\sim 0.5\%$ accuracy gain both when the full Real dataset or RBU dataset are used (for comparison, doubling the training dataset, i.e., Real$\to$RBU, yields a $+0.8\%$ performance increase). This demonstrates the effectiveness of CSD at any data scale. Additional considerations about the effectiveness of CSD are presented in the supp.~material (Sec.~\ref{sec:additional_considerations}).

\subsection{Differential decoder}
\label{subsec:diff_decoder_results}
To enhance the performance of CSD without a significant increase in GFLOPs, we introduce the differential decoder PLD-Diff (Sec.~\ref{subsec:diff}). We evaluate its effectiveness in the base configuration with 2 layers, an inner dimension of 768 and 12 attention heads and with ViT-Base encoder (Tab.~\ref{tab:architecture_details}). Tab.~\ref{tab:model_diff} shows that PLD-Diff consistently improves the performance using both Real and RBU datasets. As claimed in~\cite{ye2024differential}, in a traditional Cross-Attention mechanism, a small proportion of attention maps might focus on relevant context. Hence, this leads to poor predictions and decreases the performance. In contrast, Differential attention concentrates more on critical information, so that a performance increase can be observed. Furthermore, PLD-Diff adds $5.7$M parameters (compared to PLD-Base), but only $0.1$ GFLOPs by considering an average sequence length of 5.5.

\subsection{Comparison with State-of-the-Art}
\label{subsec:comparison_sota}
We compare our CSD technique and differential decoder with previous approaches. Specifically, we focus on three different model configurations:
\begin{itemize}
    \item \tb{CSD-S} (40.8M parameters): ViT-Small + PLD-Base
    \item \tb{CSD-B} (104.9M parameters): ViT-Base + PLD-Base
    \item \tb{CSD-D} (110.2M parameters): ViT-Base + PLD-Diff
\end{itemize}

\noindent Table~\ref{tab:sota} shows that our models outperform previous state-of-the-art models in almost all benchmarks, whether they are trained on the Real or RBU dataset. Precisely, when they are trained solely on the Real dataset, our models outperform the previous state-of-the-art models on 10 out of 11 benchmarks, while requiring significantly less parameters and GFLOPs. Our best model, CSD-D, achieves an \tb{AVG}$_{11}$ accuracy of $92.73\%$ and a \tb{wAVG}$_{6}$ accuracy of $97.42\%$, compared to CLIP4STR-L whose respective scores are $91.69\%$ and $97.04\%$. Note that our models use only $24.7\%$ of the parameters and $23.9\%$ of the GFLOPs achieved by CLIP4STR-L. By expanding the training dataset to RBU, our models continue to outperform previous models, with CSD-D achieving an \tb{AVG}$_{11}$ accuracy of $93.30\%$ and a \tb{wAVG}$_{6}$ accuracy of $97.62\%$, even outperforming the 97.42\% \tb{wAVG}$_{6}$ achieved by CLIP4STR-L when scaled to the larger RBU-Syn dataset whose size is $18$M \cite{rang2023large}. 

\noindent Even if our method also achieves similar performance compared to DTrOCR \cite{fujitake2024dtrocr} in most benchmarks, we have not included their results in this section, since they employ a training set with billions of additional images and the code/weights have not been released. 

\subsection{Additional analyses}
In the supplementary material, we report additional results and ablation studies. In Sec.~\ref{sec:gflops}, we provide a detailed analysis of GFLOPs by considering the impact of the decoder with varying sequence lengths. In Sec.~\ref{sec:csd_hparams}, we analyze the effect of CSD hyperparameters (i.e., temperature $\tau$ and KD loss mixing parameter $\alpha$). In Sec. \ref{sec:pretraining}, \ref{sec:additional_considerations} and \ref{sec:qualitative_examples} we present additional results and analyses to support the superiority of our model and methodology.

\vspace{-0.2cm}
\section{Conclusion} \label{sec:conclusion}
In this work, we present a comprehensive  analysis of encoder-decoder scaling for STR by demonstrating the significant benefits of scaling the decoder. Additionally, we introduce a novel training strategy to address label noise in real-world STR datasets. We leverage context-aware predictions generated from a teacher model through a cloze-filling approach, to distill a student model with improved performance. 
 Moreover, we propose architectural updates, including Differential Cross-Attention, to improve the effectiveness of the decoder to focus on relevant context during inference. Empirical evaluations show the superiority of our model, achieving SOTA across multiple benchmarks while using fewer parameters and reducing the computational overhead (FLOPs) compared to previous models.

{
    \small
    \bibliographystyle{ieeenat_fullname}
    \bibliography{main}

\begin{thebibliography}{54}
\providecommand{\natexlab}[1]{#1}
\providecommand{\url}[1]{\texttt{#1}}
\expandafter\ifx\csname urlstyle\endcsname\relax
  \providecommand{\doi}[1]{doi: #1}\else
  \providecommand{\doi}{doi: \begingroup \urlstyle{rm}\Url}\fi

\bibitem[Atienza(2021)]{atienza2021vision}
Rowel Atienza.
\newblock Vision transformer for fast and efficient scene text recognition.
\newblock In \emph{International conference on document analysis and recognition}, pages 319--334. Springer, 2021.

\bibitem[Baek et~al.(2019)Baek, Kim, Lee, Park, Han, Yun, Oh, and Lee]{baek2019wrong}
Jeonghun Baek, Geewook Kim, Junyeop Lee, Sungrae Park, Dongyoon Han, Sangdoo Yun, Seong~Joon Oh, and Hwalsuk Lee.
\newblock What is wrong with scene text recognition model comparisons? dataset and model analysis.
\newblock In \emph{Proceedings of the IEEE/CVF international conference on computer vision}, pages 4715--4723, 2019.

\bibitem[Bautista and Atienza(2022)]{bautista2022scene}
Darwin Bautista and Rowel Atienza.
\newblock Scene text recognition with permuted autoregressive sequence models.
\newblock In \emph{European conference on computer vision}, pages 178--196. Springer, 2022.

\bibitem[Bhunia et~al.(2021)Bhunia, Sain, Chowdhury, and Song]{bhunia2021text}
Ayan~Kumar Bhunia, Aneeshan Sain, Pinaki~Nath Chowdhury, and Yi-Zhe Song.
\newblock Text is text, no matter what: Unifying text recognition using knowledge distillation.
\newblock In \emph{Proceedings of the IEEE/CVF International Conference on Computer Vision}, pages 983--992, 2021.

\bibitem[Borisyuk et~al.(2018)Borisyuk, Gordo, and Sivakumar]{borisyuk2018rosetta}
Fedor Borisyuk, Albert Gordo, and Viswanath Sivakumar.
\newblock Rosetta: Large scale system for text detection and recognition in images.
\newblock In \emph{Proceedings of the 24th ACM SIGKDD international conference on knowledge discovery \& data mining}, pages 71--79, 2018.

\bibitem[Chng et~al.(2019)Chng, Liu, Sun, Ng, Luo, Ni, Fang, Zhang, Han, Ding, et~al.]{chng2019icdar2019}
Chee~Kheng Chng, Yuliang Liu, Yipeng Sun, Chun~Chet Ng, Canjie Luo, Zihan Ni, ChuanMing Fang, Shuaitao Zhang, Junyu Han, Errui Ding, et~al.
\newblock Icdar2019 robust reading challenge on arbitrary-shaped text-rrc-art.
\newblock In \emph{2019 International Conference on Document Analysis and Recognition (ICDAR)}, pages 1571--1576. IEEE, 2019.

\bibitem[Cubuk et~al.(2020)Cubuk, Zoph, Shlens, and Le]{cubuk2020randaugment}
Ekin~D Cubuk, Barret Zoph, Jonathon Shlens, and Quoc~V Le.
\newblock Randaugment: Practical automated data augmentation with a reduced search space.
\newblock In \emph{Proceedings of the IEEE/CVF conference on computer vision and pattern recognition workshops}, pages 702--703, 2020.

\bibitem[Deng et~al.(2009)Deng, Dong, Socher, Li, Li, and Fei-Fei]{deng2009imagenet}
Jia Deng, Wei Dong, Richard Socher, Li-Jia Li, Kai Li, and Li Fei-Fei.
\newblock Imagenet: A large-scale hierarchical image database.
\newblock In \emph{2009 IEEE conference on computer vision and pattern recognition}, pages 248--255. Ieee, 2009.

\bibitem[Dosovitskiy(2020)]{dosovitskiy2020image}
Alexey Dosovitskiy.
\newblock An image is worth 16x16 words: Transformers for image recognition at scale.
\newblock \emph{arXiv preprint arXiv:2010.11929}, 2020.

\bibitem[Fang et~al.(2021)Fang, Xie, Wang, Mao, and Zhang]{fang2021read}
Shancheng Fang, Hongtao Xie, Yuxin Wang, Zhendong Mao, and Yongdong Zhang.
\newblock Read like humans: Autonomous, bidirectional and iterative language modeling for scene text recognition.
\newblock In \emph{Proceedings of the IEEE/CVF conference on computer vision and pattern recognition}, pages 7098--7107, 2021.

\bibitem[Fujitake(2024)]{fujitake2024dtrocr}
Masato Fujitake.
\newblock Dtrocr: Decoder-only transformer for optical character recognition.
\newblock In \emph{Proceedings of the IEEE/CVF Winter Conference on Applications of Computer Vision}, pages 8025--8035, 2024.

\bibitem[Graves et~al.(2006)Graves, Fern{\'a}ndez, Gomez, and Schmidhuber]{graves2006connectionist}
Alex Graves, Santiago Fern{\'a}ndez, Faustino Gomez, and J{\"u}rgen Schmidhuber.
\newblock Connectionist temporal classification: labelling unsegmented sequence data with recurrent neural networks.
\newblock In \emph{Proceedings of the 23rd international conference on Machine learning}, pages 369--376, 2006.

\bibitem[He et~al.(2016)He, Huang, Qiao, Loy, and Tang]{he2016reading}
Pan He, Weilin Huang, Yu Qiao, Chen Loy, and Xiaoou Tang.
\newblock Reading scene text in deep convolutional sequences.
\newblock In \emph{Proceedings of the AAAI conference on artificial intelligence}, 2016.

\bibitem[Hinton(2015)]{hinton2015distilling}
Geoffrey Hinton.
\newblock Distilling the knowledge in a neural network.
\newblock \emph{arXiv preprint arXiv:1503.02531}, 2015.

\bibitem[Jiang et~al.(2023)Jiang, Wang, Peng, Liu, and Jin]{jiang2023revisiting}
Qing Jiang, Jiapeng Wang, Dezhi Peng, Chongyu Liu, and Lianwen Jin.
\newblock Revisiting scene text recognition: A data perspective.
\newblock In \emph{Proceedings of the IEEE/CVF international conference on computer vision}, pages 20543--20554, 2023.

\bibitem[Kaplan et~al.(2020)Kaplan, McCandlish, Henighan, Brown, Chess, Child, Gray, Radford, Wu, and Amodei]{kaplan2020scaling}
Jared Kaplan, Sam McCandlish, Tom Henighan, Tom~B Brown, Benjamin Chess, Rewon Child, Scott Gray, Alec Radford, Jeffrey Wu, and Dario Amodei.
\newblock Scaling laws for neural language models.
\newblock \emph{arXiv preprint arXiv:2001.08361}, 2020.

\bibitem[Karatzas et~al.(2013)Karatzas, Shafait, Uchida, Iwamura, i~Bigorda, Mestre, Mas, Mota, Almazan, and De~Las~Heras]{karatzas2013icdar}
Dimosthenis Karatzas, Faisal Shafait, Seiichi Uchida, Masakazu Iwamura, Lluis~Gomez i Bigorda, Sergi~Robles Mestre, Joan Mas, David~Fernandez Mota, Jon~Almazan Almazan, and Lluis~Pere De~Las~Heras.
\newblock Icdar 2013 robust reading competition.
\newblock In \emph{2013 12th international conference on document analysis and recognition}, pages 1484--1493. IEEE, 2013.

\bibitem[Karatzas et~al.(2015)Karatzas, Gomez-Bigorda, Nicolaou, Ghosh, Bagdanov, Iwamura, Matas, Neumann, Chandrasekhar, Lu, et~al.]{karatzas2015icdar}
Dimosthenis Karatzas, Lluis Gomez-Bigorda, Anguelos Nicolaou, Suman Ghosh, Andrew Bagdanov, Masakazu Iwamura, Jiri Matas, Lukas Neumann, Vijay~Ramaseshan Chandrasekhar, Shijian Lu, et~al.
\newblock Icdar 2015 competition on robust reading.
\newblock In \emph{2015 13th international conference on document analysis and recognition (ICDAR)}, pages 1156--1160. IEEE, 2015.

\bibitem[Krylov et~al.(2021)Krylov, Nosov, and Sovrasov]{krylov2021open}
Ilya Krylov, Sergei Nosov, and Vladislav Sovrasov.
\newblock Open images v5 text annotation and yet another mask text spotter.
\newblock In \emph{Asian Conference on Machine Learning}, pages 379--389. PMLR, 2021.

\bibitem[Lee and Osindero(2016)]{lee2016recursive}
Chen-Yu Lee and Simon Osindero.
\newblock Recursive recurrent nets with attention modeling for ocr in the wild.
\newblock In \emph{Proceedings of the IEEE conference on computer vision and pattern recognition}, pages 2231--2239, 2016.

\bibitem[Li et~al.(2023)Li, Lv, Chen, Cui, Lu, Florencio, Zhang, Li, and Wei]{li2023trocr}
Minghao Li, Tengchao Lv, Jingye Chen, Lei Cui, Yijuan Lu, Dinei Florencio, Cha Zhang, Zhoujun Li, and Furu Wei.
\newblock Trocr: Transformer-based optical character recognition with pre-trained models.
\newblock In \emph{Proceedings of the AAAI Conference on Artificial Intelligence}, pages 13094--13102, 2023.

\bibitem[Liu et~al.(2016)Liu, Chen, Wong, Su, and Han]{liu2016star}
Wei Liu, Chaofeng Chen, Kwan-Yee~K Wong, Zhizhong Su, and Junyu Han.
\newblock Star-net: a spatial attention residue network for scene text recognition.
\newblock In \emph{BMVC}, page~7, 2016.

\bibitem[Long et~al.(2021)Long, He, and Yao]{long2021scene}
Shangbang Long, Xin He, and Cong Yao.
\newblock Scene text detection and recognition: The deep learning era.
\newblock \emph{International Journal of Computer Vision}, 129\penalty0 (1):\penalty0 161--184, 2021.

\bibitem[Loshchilov(2017)]{loshchilov2017decoupled}
I Loshchilov.
\newblock Decoupled weight decay regularization.
\newblock \emph{arXiv preprint arXiv:1711.05101}, 2017.

\bibitem[Mishra et~al.(2012)Mishra, Alahari, and Jawahar]{mishra2012scene}
Anand Mishra, Karteek Alahari, and CV Jawahar.
\newblock Scene text recognition using higher order language priors.
\newblock In \emph{BMVC-British machine vision conference}. BMVA, 2012.

\bibitem[Nayef et~al.(2019)Nayef, Patel, Busta, Chowdhury, Karatzas, Khlif, Matas, Pal, Burie, Liu, et~al.]{nayef2019icdar2019}
Nibal Nayef, Yash Patel, Michal Busta, Pinaki~Nath Chowdhury, Dimosthenis Karatzas, Wafa Khlif, Jiri Matas, Umapada Pal, Jean-Christophe Burie, Cheng-lin Liu, et~al.
\newblock Icdar2019 robust reading challenge on multi-lingual scene text detection and recognition—rrc-mlt-2019.
\newblock In \emph{2019 International conference on document analysis and recognition (ICDAR)}, pages 1582--1587. IEEE, 2019.

\bibitem[Ouali et~al.(2022)Ouali, Halima, and Wali]{ouali2022text}
Imene Ouali, Mohamed~Ben Halima, and Ali Wali.
\newblock Text detection and recognition using augmented reality and deep learning.
\newblock In \emph{International conference on advanced information networking and applications}, pages 13--23. Springer, 2022.

\bibitem[Phan et~al.(2013)Phan, Shivakumara, Tian, and Tan]{phan2013recognizing}
Trung~Quy Phan, Palaiahnakote Shivakumara, Shangxuan Tian, and Chew~Lim Tan.
\newblock Recognizing text with perspective distortion in natural scenes.
\newblock In \emph{Proceedings of the IEEE international conference on computer vision}, pages 569--576, 2013.

\bibitem[Radford et~al.(2019)Radford, Wu, Child, Luan, Amodei, Sutskever, et~al.]{radford2019language}
Alec Radford, Jeffrey Wu, Rewon Child, David Luan, Dario Amodei, Ilya Sutskever, et~al.
\newblock Language models are unsupervised multitask learners.
\newblock \emph{OpenAI blog}, 1\penalty0 (8):\penalty0 9, 2019.

\bibitem[Radford et~al.(2021)Radford, Kim, Hallacy, Ramesh, Goh, Agarwal, Sastry, Askell, Mishkin, Clark, et~al.]{radford2021learning}
Alec Radford, Jong~Wook Kim, Chris Hallacy, Aditya Ramesh, Gabriel Goh, Sandhini Agarwal, Girish Sastry, Amanda Askell, Pamela Mishkin, Jack Clark, et~al.
\newblock Learning transferable visual models from natural language supervision.
\newblock In \emph{International conference on machine learning}, pages 8748--8763. PMLR, 2021.

\bibitem[Rang et~al.(2023)Rang, Bi, Liu, Wang, and Han]{rang2023large}
Miao Rang, Zhenni Bi, Chuanjian Liu, Yunhe Wang, and Kai Han.
\newblock Large ocr model: An empirical study of scaling law for ocr.
\newblock \emph{arXiv preprint arXiv:2401.00028}, 2023.

\bibitem[Risnumawan et~al.(2014)Risnumawan, Shivakumara, Chan, and Tan]{risnumawan2014robust}
Anhar Risnumawan, Palaiahankote Shivakumara, Chee~Seng Chan, and Chew~Lim Tan.
\newblock A robust arbitrary text detection system for natural scene images.
\newblock \emph{Expert Systems with Applications}, 41\penalty0 (18):\penalty0 8027--8048, 2014.

\bibitem[Shazeer(2020)]{shazeer2020glu}
Noam Shazeer.
\newblock Glu variants improve transformer.
\newblock \emph{arXiv preprint arXiv:2002.05202}, 2020.

\bibitem[Sheng et~al.(2019)Sheng, Chen, and Xu]{sheng2019nrtr}
Fenfen Sheng, Zhineng Chen, and Bo Xu.
\newblock Nrtr: A no-recurrence sequence-to-sequence model for scene text recognition.
\newblock In \emph{2019 International conference on document analysis and recognition (ICDAR)}, pages 781--786. IEEE, 2019.

\bibitem[Shi et~al.(2016{\natexlab{a}})Shi, Bai, and Yao]{shi2016end}
Baoguang Shi, Xiang Bai, and Cong Yao.
\newblock An end-to-end trainable neural network for image-based sequence recognition and its application to scene text recognition.
\newblock \emph{IEEE transactions on pattern analysis and machine intelligence}, 39\penalty0 (11):\penalty0 2298--2304, 2016{\natexlab{a}}.

\bibitem[Shi et~al.(2016{\natexlab{b}})Shi, Wang, Lyu, Yao, and Bai]{shi2016robust}
Baoguang Shi, Xinggang Wang, Pengyuan Lyu, Cong Yao, and Xiang Bai.
\newblock Robust scene text recognition with automatic rectification.
\newblock In \emph{Proceedings of the IEEE conference on computer vision and pattern recognition}, pages 4168--4176, 2016{\natexlab{b}}.

\bibitem[Shi et~al.(2017)Shi, Yao, Liao, Yang, Xu, Cui, Belongie, Lu, and Bai]{shi2017icdar2017}
Baoguang Shi, Cong Yao, Minghui Liao, Mingkun Yang, Pei Xu, Linyan Cui, Serge Belongie, Shijian Lu, and Xiang Bai.
\newblock Icdar2017 competition on reading chinese text in the wild (rctw-17).
\newblock In \emph{2017 14th iapr international conference on document analysis and recognition (ICDAR)}, pages 1429--1434. IEEE, 2017.

\bibitem[Shi et~al.(2018)Shi, Yang, Wang, Lyu, Yao, and Bai]{shi2018aster}
Baoguang Shi, Mingkun Yang, Xinggang Wang, Pengyuan Lyu, Cong Yao, and Xiang Bai.
\newblock Aster: An attentional scene text recognizer with flexible rectification.
\newblock \emph{IEEE transactions on pattern analysis and machine intelligence}, 41\penalty0 (9):\penalty0 2035--2048, 2018.

\bibitem[Singh et~al.(2021)Singh, Pang, Toh, Huang, Galuba, and Hassner]{singh2021textocr}
Amanpreet Singh, Guan Pang, Mandy Toh, Jing Huang, Wojciech Galuba, and Tal Hassner.
\newblock Textocr: Towards large-scale end-to-end reasoning for arbitrary-shaped scene text.
\newblock In \emph{Proceedings of the IEEE/CVF conference on computer vision and pattern recognition}, pages 8802--8812, 2021.

\bibitem[Smith and Topin(2019)]{smith2019super}
Leslie~N Smith and Nicholay Topin.
\newblock Super-convergence: Very fast training of neural networks using large learning rates.
\newblock In \emph{Artificial intelligence and machine learning for multi-domain operations applications}, pages 369--386. SPIE, 2019.

\bibitem[Su et~al.(2024)Su, Ahmed, Lu, Pan, Bo, and Liu]{su2024roformer}
Jianlin Su, Murtadha Ahmed, Yu Lu, Shengfeng Pan, Wen Bo, and Yunfeng Liu.
\newblock Roformer: Enhanced transformer with rotary position embedding.
\newblock \emph{Neurocomputing}, 568:\penalty0 127063, 2024.

\bibitem[Sun et~al.(2019)Sun, Ni, Chng, Liu, Luo, Ng, Han, Ding, Liu, Karatzas, et~al.]{sun2019icdar}
Yipeng Sun, Zihan Ni, Chee-Kheng Chng, Yuliang Liu, Canjie Luo, Chun~Chet Ng, Junyu Han, Errui Ding, Jingtuo Liu, Dimosthenis Karatzas, et~al.
\newblock Icdar 2019 competition on large-scale street view text with partial labeling-rrc-lsvt.
\newblock In \emph{2019 International Conference on Document Analysis and Recognition (ICDAR)}, pages 1557--1562. IEEE, 2019.

\bibitem[Vaidya et~al.(2023)Vaidya, Sharma, Gatti, and Mishra]{vaidya2023show}
Shreyas Vaidya, Arvind~Kumar Sharma, Prajwal Gatti, and Anand Mishra.
\newblock Show me the world in my language: Establishing the first baseline for scene-text to scene-text translation.
\newblock \emph{arXiv preprint arXiv:2308.03024}, 2023.

\bibitem[Veit et~al.(2016)Veit, Matera, Neumann, Matas, and Belongie]{veit2016coco}
Andreas Veit, Tomas Matera, Lukas Neumann, Jiri Matas, and Serge Belongie.
\newblock Coco-text: Dataset and benchmark for text detection and recognition in natural images.
\newblock \emph{arXiv preprint arXiv:1601.07140}, 2016.

\bibitem[Wang et~al.(2011)Wang, Babenko, and Belongie]{wang2011end}
Kai Wang, Boris Babenko, and Serge Belongie.
\newblock End-to-end scene text recognition.
\newblock In \emph{2011 International conference on computer vision}, pages 1457--1464. IEEE, 2011.

\bibitem[Wang et~al.(2020)Wang, Zhu, Jin, Luo, Chen, Wu, Wang, and Cai]{wang2020decoupled}
Tianwei Wang, Yuanzhi Zhu, Lianwen Jin, Canjie Luo, Xiaoxue Chen, Yaqiang Wu, Qianying Wang, and Mingxiang Cai.
\newblock Decoupled attention network for text recognition.
\newblock In \emph{Proceedings of the AAAI conference on artificial intelligence}, pages 12216--12224, 2020.

\bibitem[Wang et~al.(2021)Wang, Xie, Fang, Wang, Zhu, and Zhang]{wang2021two}
Yuxin Wang, Hongtao Xie, Shancheng Fang, Jing Wang, Shenggao Zhu, and Yongdong Zhang.
\newblock From two to one: A new scene text recognizer with visual language modeling network.
\newblock In \emph{Proceedings of the IEEE/CVF International Conference on Computer Vision}, pages 14194--14203, 2021.

\bibitem[Wang et~al.(2023)Wang, Xie, Wang, Xu, Zhang, and Zhang]{wang2023symmetrical}
Zixiao Wang, Hongtao Xie, Yuxin Wang, Jianjun Xu, Boqiang Zhang, and Yongdong Zhang.
\newblock Symmetrical linguistic feature distillation with clip for scene text recognition.
\newblock In \emph{Proceedings of the 31st ACM International Conference on Multimedia}, pages 509--518, 2023.

\bibitem[Yang(2019)]{yang2019xlnet}
Zhilin Yang.
\newblock Xlnet: Generalized autoregressive pretraining for language understanding.
\newblock \emph{arXiv preprint arXiv:1906.08237}, 2019.

\bibitem[Ye et~al.(2024)Ye, Dong, Xia, Sun, Zhu, Huang, and Wei]{ye2024differential}
Tianzhu Ye, Li Dong, Yuqing Xia, Yutao Sun, Yi Zhu, Gao Huang, and Furu Wei.
\newblock Differential transformer.
\newblock \emph{arXiv preprint arXiv:2410.05258}, 2024.

\bibitem[Zhang et~al.(2020)Zhang, Ding, Peng, Fu, and Wang]{zhang2020street}
Chongsheng Zhang, Weiping Ding, Guowen Peng, Feifei Fu, and Wei Wang.
\newblock Street view text recognition with deep learning for urban scene understanding in intelligent transportation systems.
\newblock \emph{IEEE Transactions on Intelligent Transportation Systems}, 22\penalty0 (7):\penalty0 4727--4743, 2020.

\bibitem[Zhang et~al.(2019)Zhang, Zhou, Jiang, Song, Li, Zhou, Wang, Wang, Liao, Yang, et~al.]{zhang2019icdar}
Rui Zhang, Yongsheng Zhou, Qianyi Jiang, Qi Song, Nan Li, Kai Zhou, Lei Wang, Dong Wang, Minghui Liao, Mingkun Yang, et~al.
\newblock Icdar 2019 robust reading challenge on reading chinese text on signboard.
\newblock In \emph{2019 international conference on document analysis and recognition (ICDAR)}, pages 1577--1581. IEEE, 2019.

\bibitem[Zhang et~al.(2017)Zhang, Gueguen, Zharkov, Zhang, Seifert, and Kadlec]{zhang2017uber}
Ying Zhang, Lionel Gueguen, Ilya Zharkov, Peter Zhang, Keith Seifert, and Ben Kadlec.
\newblock Uber-text: A large-scale dataset for optical character recognition from street-level imagery.
\newblock In \emph{SUNw: Scene Understanding Workshop-CVPR}, page~5, 2017.

\bibitem[Zhao et~al.(2023)Zhao, Quan, Zhu, and Yang]{zhao2023clip4str}
Shuai Zhao, Ruijie Quan, Linchao Zhu, and Yi Yang.
\newblock Clip4str: A simple baseline for scene text recognition with pre-trained vision-language model.
\newblock \emph{arXiv preprint arXiv:2305.14014}, 2023.

\end{thebibliography}
}

\clearpage
\setcounter{page}{1}
\maketitlesupplementary

\section{Context Update in Permutation Language Decoder}
\label{sec:context_update}

\begin{figure}[b!]
\centering
\includegraphics[width=0.99\linewidth]{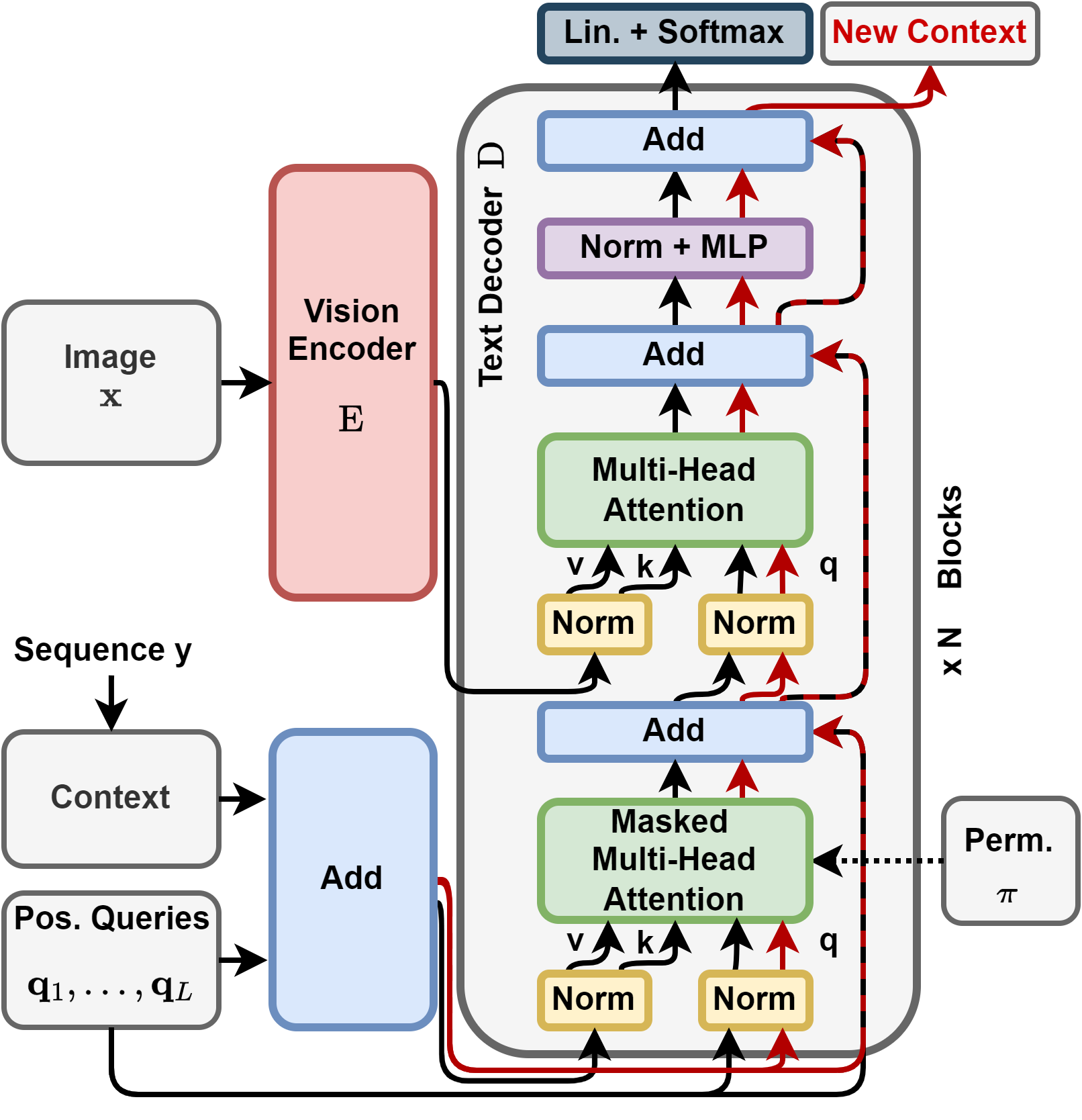}
\caption{\tb{Diagram of the context update in PLD.} The positional queries are updated following the black arrows (as in our Permutation Language Decoder). The context is updated following the red arrows: it is used as input of the query stream in a second forward pass before using it in the following block.}
\label{fig:context_update}
\end{figure}

In Sec.~\ref{sec:method}, we introduced the architecture of the Permutation Language Decoder (PLD) used in our STR model. Specifically, in our implementation, each block of PLD receives the output of the previous block as the input of the query stream, while the key-value stream is provided with the same context and vision tokens across all blocks. It simplifies the original approach used in PARSeq~\cite{bautista2022scene}, which updates the context when multiple blocks are presented.  
While the positional queries in PARSeq follow the same query stream as in our implementation, PARSeq additionally provides the context as input to the query stream in a second forward pass. This is done in order to update it before using it as input of the key-value stream of the next block.
Fig.~\ref{fig:context_update} shows the diagram of PLD in the PARSeq implementation: the positional queries follow the same path as in our implementation (black arrows), the context is updated following the red arrows.

\noindent Empirically, we found that this additional context update degrades the performance. Considering the average word accuracy across 11 benchmarks (\tb{AVG}$_{11}$), the performance of ViT-Base with PLD-Base decreases $0.15\%$, while ViT-Small and PLD-Base have a decrease of $0.19\%$. Moreover, since the context is also updated, the computational complexity is also increased. To this end, in all our analyses and experiments, we do not update the context as a default setting.

\noindent \textbf{Remark.} In PARSeq paper, they present the results using a single-block decoder so the context is actually not updated. However, their official implementation updates the context when multiple blocks are used.

\begin{figure}[t!]
\centering
\includegraphics[width=0.99\linewidth]{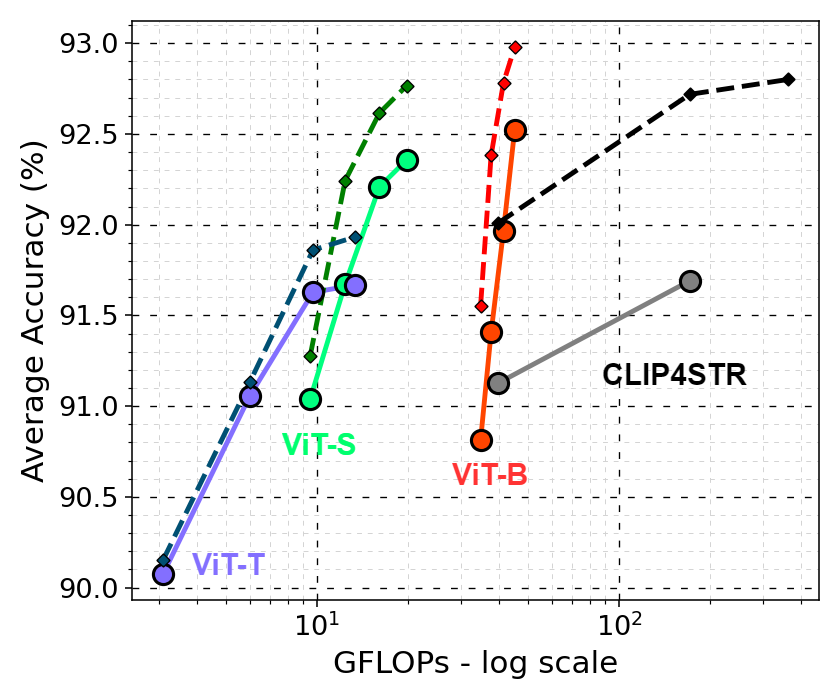}
\caption{\textbf{Average word accuracy} (\%) on $11$ STR benchmarks for the models with ViT-T, ViT-S and ViT-B vision encoders and $4$ different decoder sizes (see Sec. \ref{subsec:scaling_analysis}). Results are compared with the previous state-of-the-art model, CLIP4STR~\cite{zhao2023clip4str}. Results using \tb{Real} training dataset (3.3M images) are depicted with solid lines and circle markers, while results using \tb{RBU} training dataset (6.5M images) are shown with dashed lines and diamond markers. The x-axis represents the \tb{total number GFLOPs} on a logarithmic scale.}
\vspace{-0.7cm}
\label{fig:flops_vs_accuracy}
\end{figure}

\section{Computational efficiency}
\label{sec:gflops}
In our STR model, the encoder presents a fixed computational cost, as it processes the vision tokens in a single forward pass. In contrast, the computational cost of the decoder depends on the sequence length due to the use of auto-regressive (AR) decoding, which has been shown to outperform non-autoregressive (NAR) methods \cite{bautista2022scene}. In Sec.~\ref{subsec:scaling_results}, we demonstrated that increasing the decoder size is effective to improve performance. In this section, we analyze the impact of decoder size on overall GFLOPs.

\begin{figure}[t!]
\centering
\includegraphics[width=0.99\linewidth]{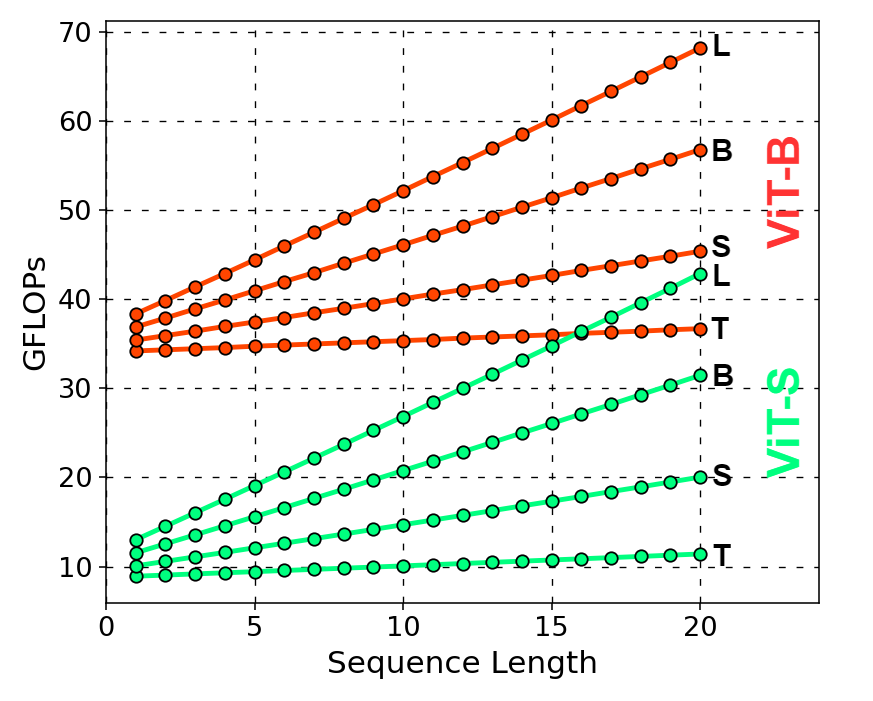}
\caption{\tb{GFLOPs for different sequence lengths.} The x-axis represents the sequence length (from 1 to 20 characters), while the y-axis represents the number of GFLOPs. Results are reported for ViT-Base and ViT-Small encoders paired with different decoders (PLD-T, PLD-S, PLD-B and PLD-L).}
\vspace{-0.7cm}
\label{fig:flops_seq_length}
\end{figure}

\noindent Fig.~\ref{fig:flops_vs_accuracy} illustrates how the average model accuracies and the GLOPs change together. A similar plot is provided in Fig.~\ref{fig:plot_scaling} for the average model accuracy and the total number of parameters.
The GFLOPs are calculated based on the average sequence length of $5.5$, which corresponds to the average sequence length across all benchmark datasets. The plot reveals a similar trend that is observed for the number of parameters. Additionally, Fig.~\ref{fig:flops_seq_length} shows how GFLOPs vary across different sequence lengths (from 3 to 20 characters) for various decoder sizes using ViT-B and ViT-S as encoders. Notably, for short sequence lengths, the encoder has the highest computational cost compared to the decoder. However, as the sequence length increases, decoder's GFLOPs increase, particularly for larger decoders. In most STR tasks, efficiency for long sequences is not a primary target since this kind of sequences is less common in natural scene settings.

\noindent  \tb{Remark.} When referring to the sequence length, we specifically consider the number of characters to be decoded. In the actual implementation, two additional special tokens are also decoded: the beginning-of-sequence token (BOS) and the end-of-sequence token (EOS), which mark the start and end of decoding process, respectively. The computation of these tokens is included in the GFLOPs calculation for any sequence length.

\section{Cloze Self-Distillation: hyperparameters}
\label{sec:csd_hparams}

In Sec. \ref{subsec:csd}, we introduced Cloze Self-Distillation, our novel technique to train STR models on real data. The objective of CSD is presented in Eq. \ref{eq:CSD_objective} that we report here for convenience:

\begin{equation*}
    \min_{\theta}  \mathbb{E}_{\substack{(\mb{x}, \mb{y}) \sim \mathcal{D} \\ \boldsymbol{\pi} \sim \Pi \\ t \sim [1, L]}} \left[ - \log p_\theta(y_{\pi_t}|\mb{y}_{\boldsymbol{\pi}_{<t}}, \mb{x}) + \alpha\text{KD}_{\boldsymbol{\pi}, t}(\mb{x}, \mb{y})  \right] 
\end{equation*}

\noindent In the experiments presented in the main text, we set the mixing hyperparameter and distillation temperature to $\alpha = 0.1$ and $\tau = 2.0$, respectively. In this section, we present a post-hoc ablation study to show that CSD is not highly sensitive to these hyperparameters. To provide consistent results for different values of $\alpha$ without changing the learning rate and training dynamics, in this section, we multiply the loss by $\frac{1 + \alpha_0}{1 + \alpha}$, where $\alpha_0 = 0.1$ is our base value for $\alpha$. 

\begin{table}[t!]
  \centering
  \begin{tabular}{lccc}
    \toprule
    & $\boldsymbol\alpha \mb{= 0.1}$ & $\boldsymbol\alpha \mb{= 0.5}$ & $\boldsymbol\alpha \mb{= 1.0}$ \\
    \midrule
     $\boldsymbol\tau \mb{= 1.0}$ & 92.4 & 92.4 & 92.5 \\
     $\boldsymbol\tau \mb{= 2.0}$ & 92.5 & 92.5 & 92.5 \\
     $\boldsymbol\tau \mb{= 3.0}$ & 92.5 & \tb{92.6} & \tb{92.6} \\

    \bottomrule
  \end{tabular}
  \caption{\tb{CSD hyperpameters.} Average word accuracy (\%) \tb{AVG}$_{11}$ using CSD-B (ViT-Base + PLD-Base) with Real dataset for different values of mixing parameter $\alpha$ and temperature $\tau$.}
  \vspace{-0.5cm}
  \label{tab:csd_hyperparameters}
\end{table}

\noindent Tab.~\ref{tab:csd_hyperparameters} shows that for each combination of $\alpha$ and $\tau$ within the considered range, the average word accuracy of CSD surpasses both the accuracy achieved using solely pseudolabels ($92.3\%$) and the accuracy obtained through conventional training methods ($92.0\%$). Moreover, increasing the temperature and mixing parameter appears to further enhance performance beyond the results presented in the main text.

\section{Architecture Analysis}
\label{sec:pretraining}

\begin{table*}[b!]
\centering
\resizebox{0.97\linewidth}{!}{
\begin{tabular}{cccccc}
\addtolength{\tabcolsep}{-0.3em}
\multirow{2}{*}[0.92in]{\includegraphics[width= 0.3\linewidth, height=0.31\linewidth]{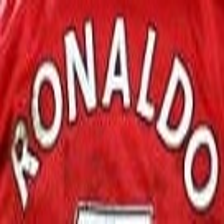}} &  \raisebox{3\normalbaselineskip}[0pt][0pt]{\rotatebox[origin=c]{90}{\small \tb{STANDARD}}} & \includegraphics[width=0.15\linewidth, height=0.15\linewidth]{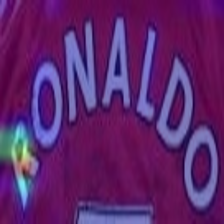}& \includegraphics[width=0.15\linewidth, height=0.15\linewidth]{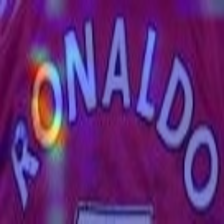}& \includegraphics[width=0.15\linewidth, height=0.15\linewidth]{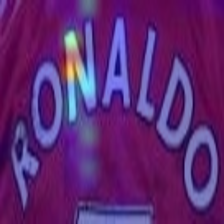}& \includegraphics[width=0.15\linewidth, height=0.15\linewidth]{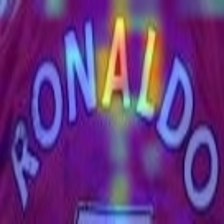}\\ &\raisebox{3\normalbaselineskip}[0pt][0pt]{\rotatebox[origin=c]{90}{\small  \tb{DIFF.}}}&\includegraphics[width=0.15\linewidth,  height=0.15\linewidth]{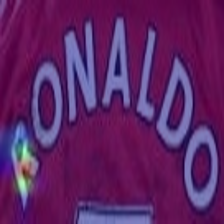}& \includegraphics[width=0.15\linewidth, height=0.15\linewidth]{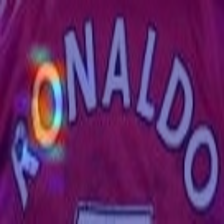}& \includegraphics[width=0.15\linewidth, height=0.15\linewidth]{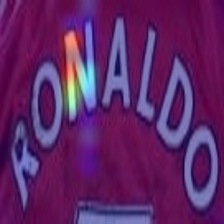} & \includegraphics[width=0.15\linewidth, height=0.15\linewidth]{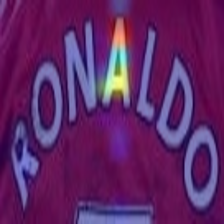}\\[10ex]

\multirow{2}{*}[0.92in]{\includegraphics[width= 0.3\linewidth, height=0.31\linewidth]{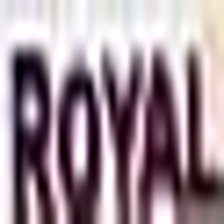}} &  \raisebox{3\normalbaselineskip}[0pt][0pt]{\rotatebox[origin=c]{90}{\small \tb{STANDARD}}} & \includegraphics[width=0.15\linewidth, height=0.15\linewidth]{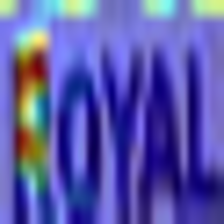}& \includegraphics[width=0.15\linewidth, height=0.15\linewidth]{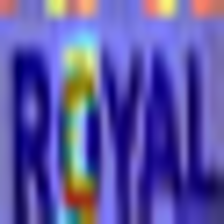}& \includegraphics[width=0.15\linewidth, height=0.15\linewidth]{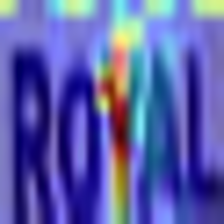}& \includegraphics[width=0.15\linewidth, height=0.15\linewidth]{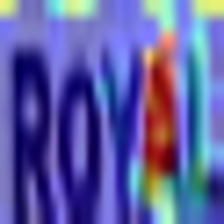}\\ &\raisebox{3\normalbaselineskip}[0pt][0pt]{\rotatebox[origin=c]{90}{\small  \tb{DIFF.}}}&\includegraphics[width=0.15\linewidth,  height=0.15\linewidth]{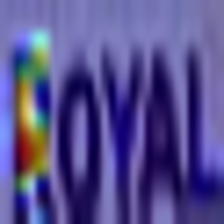}& \includegraphics[width=0.15\linewidth, height=0.15\linewidth]{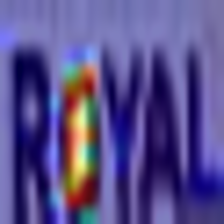}& \includegraphics[width=0.15\linewidth, height=0.15\linewidth]{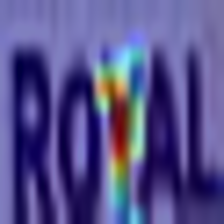} & \includegraphics[width=0.15\linewidth, height=0.15\linewidth]{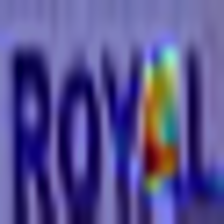}\\[10ex]

\multirow{2}{*}[0.92in]{\includegraphics[width= 0.3\linewidth, height=0.31\linewidth]{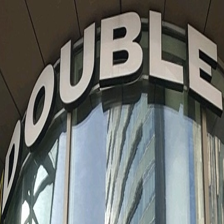}} &  \raisebox{3\normalbaselineskip}[0pt][0pt]{\rotatebox[origin=c]{90}{\small \tb{STANDARD}}} & \includegraphics[width=0.15\linewidth, height=0.15\linewidth]{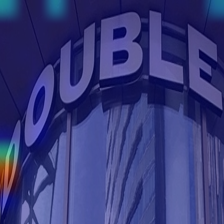}& \includegraphics[width=0.15\linewidth, height=0.15\linewidth]{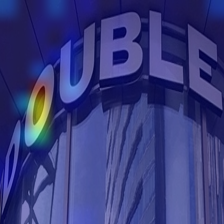}& \includegraphics[width=0.15\linewidth, height=0.15\linewidth]{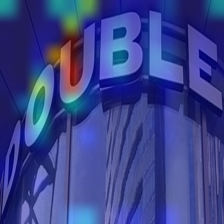}& \includegraphics[width=0.15\linewidth, height=0.15\linewidth]{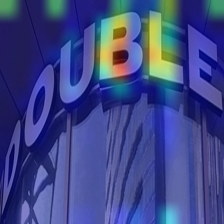}\\ &\raisebox{3\normalbaselineskip}[0pt][0pt]{\rotatebox[origin=c]{90}{\small  \tb{DIFF.}}}&\includegraphics[width=0.15\linewidth,  height=0.15\linewidth]{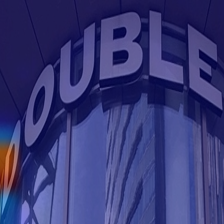}& \includegraphics[width=0.15\linewidth, height=0.15\linewidth]{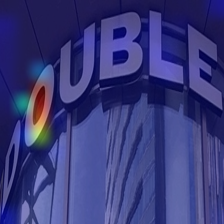}& \includegraphics[width=0.15\linewidth, height=0.15\linewidth]{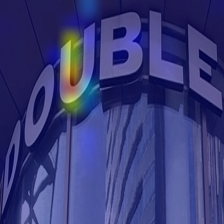} & \includegraphics[width=0.15\linewidth, height=0.15\linewidth]{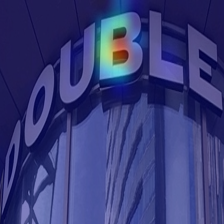}\\[10ex]
\end{tabular}}
\caption{\tb{Comparison of Attention Maps.} Attention maps of the last Cross-Attention in the last block of the Permutation Language Decoder. On the left: the original input image. First row of each section: attention maps obtained with the standard Cross-Attention. Second row of each section: attention maps obtained with our Differential Cross-Attention.}
\label{tab:maps1}
\end{table*}

\begin{table*}[t!]
\centering
\resizebox{0.99\linewidth}{!}{
\begin{tabular}{cccccc}
\addtolength{\tabcolsep}{-0.3em}
\multirow{2}{*}[0.92in]{\includegraphics[width= 0.3\linewidth, height=0.31\linewidth]{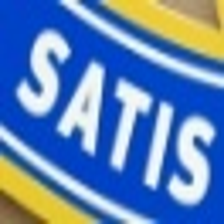}} &  \raisebox{3\normalbaselineskip}[0pt][0pt]{\rotatebox[origin=c]{90}{\small \tb{STANDARD}}} & \includegraphics[width=0.15\linewidth, height=0.15\linewidth]{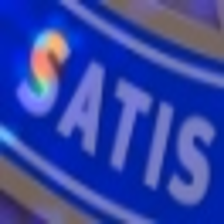}& \includegraphics[width=0.15\linewidth, height=0.15\linewidth]{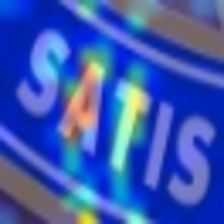}& \includegraphics[width=0.15\linewidth, height=0.15\linewidth]{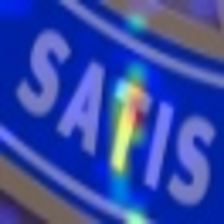}& \includegraphics[width=0.15\linewidth, height=0.15\linewidth]{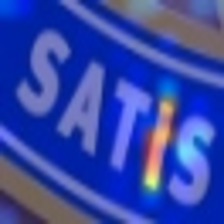}\\ &\raisebox{3\normalbaselineskip}[0pt][0pt]{\rotatebox[origin=c]{90}{\small  \tb{DIFF.}}}&\includegraphics[width=0.15\linewidth,  height=0.15\linewidth]{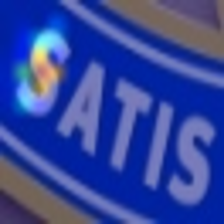}& \includegraphics[width=0.15\linewidth, height=0.15\linewidth]{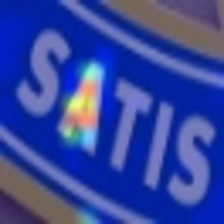}& \includegraphics[width=0.15\linewidth, height=0.15\linewidth]{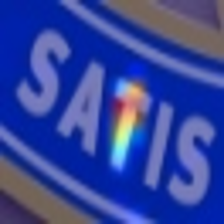} & \includegraphics[width=0.15\linewidth, height=0.15\linewidth]{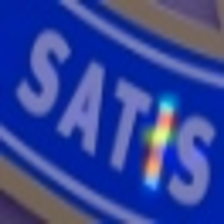}\\[10ex]

\multirow{2}{*}[0.92in]{\includegraphics[width= 0.3\linewidth, height=0.31\linewidth]{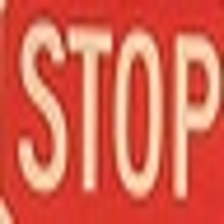}} &  \raisebox{3\normalbaselineskip}[0pt][0pt]{\rotatebox[origin=c]{90}{\small \tb{STANDARD}}} & \includegraphics[width=0.15\linewidth, height=0.15\linewidth]{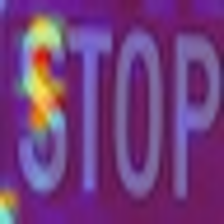}& \includegraphics[width=0.15\linewidth, height=0.15\linewidth]{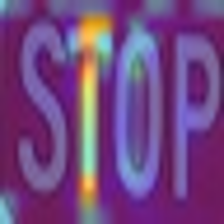}& \includegraphics[width=0.15\linewidth, height=0.15\linewidth]{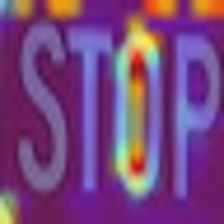}& \includegraphics[width=0.15\linewidth, height=0.15\linewidth]{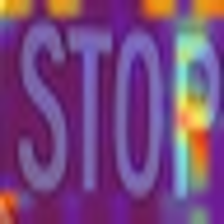}\\ &\raisebox{3\normalbaselineskip}[0pt][0pt]{\rotatebox[origin=c]{90}{\small  \tb{DIFF.}}}&\includegraphics[width=0.15\linewidth,  height=0.15\linewidth]{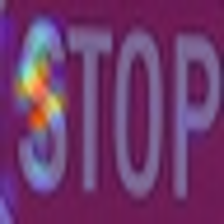}& \includegraphics[width=0.15\linewidth, height=0.15\linewidth]{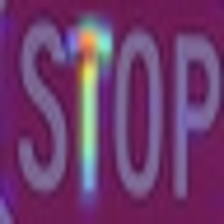}& \includegraphics[width=0.15\linewidth, height=0.15\linewidth]{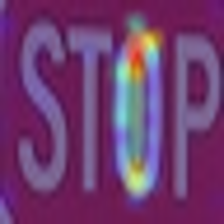} & \includegraphics[width=0.15\linewidth, height=0.15\linewidth]{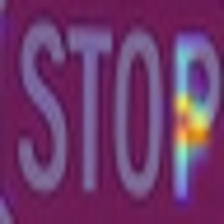}\\[10ex]

\multirow{2}{*}[0.92in]{\includegraphics[width= 0.3\linewidth, height=0.31\linewidth]{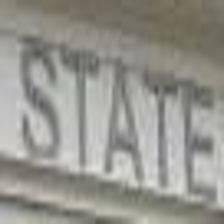}} &  \raisebox{3\normalbaselineskip}[0pt][0pt]{\rotatebox[origin=c]{90}{\small \tb{STANDARD}}} & \includegraphics[width=0.15\linewidth, height=0.15\linewidth]{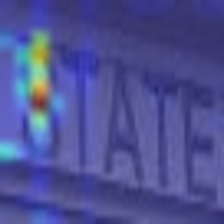}& \includegraphics[width=0.15\linewidth, height=0.15\linewidth]{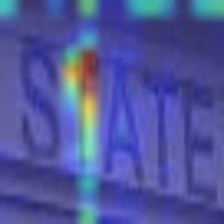}& \includegraphics[width=0.15\linewidth, height=0.15\linewidth]{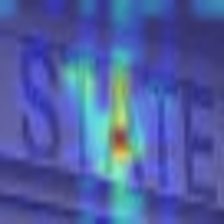}& \includegraphics[width=0.15\linewidth, height=0.15\linewidth]{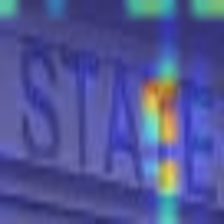}\\ &\raisebox{3\normalbaselineskip}[0pt][0pt]{\rotatebox[origin=c]{90}{\small  \tb{DIFF.}}}&\includegraphics[width=0.15\linewidth,  height=0.15\linewidth]{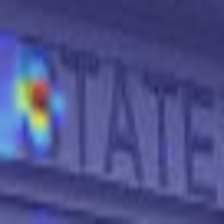}& \includegraphics[width=0.15\linewidth, height=0.15\linewidth]{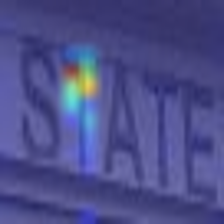}& \includegraphics[width=0.15\linewidth, height=0.15\linewidth]{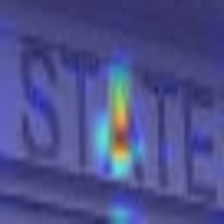} & \includegraphics[width=0.15\linewidth, height=0.15\linewidth]{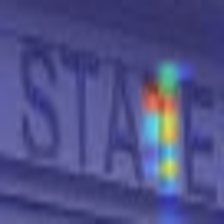}\\[10ex]
\end{tabular}}
\caption{\tb{Comparison of Attention Maps.} Attention maps of the last Cross-Attention in the last block of the Permutation Language Decoder. On the left: the original input image. First row of each section: attention maps obtained with the standard Cross-Attention. Second row of each section: attention maps obtained with our Differential Cross-Attention.}
\label{tab:maps2}
\end{table*}

\begin{table*}[t!]
\centering
\resizebox{0.99\linewidth}{!}{
\begin{tabular}{cccccc}
\addtolength{\tabcolsep}{-0.3em}
\multirow{2}{*}[0.92in]{\includegraphics[width= 0.3\linewidth, height=0.31\linewidth]{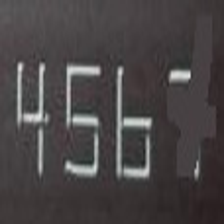}} &  \raisebox{3\normalbaselineskip}[0pt][0pt]{\rotatebox[origin=c]{90}{\small \tb{STANDARD}}} & \includegraphics[width=0.15\linewidth, height=0.15\linewidth]{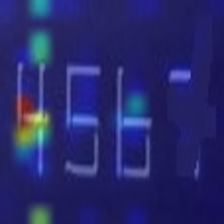}& \includegraphics[width=0.15\linewidth, height=0.15\linewidth]{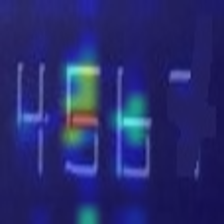}& \includegraphics[width=0.15\linewidth, height=0.15\linewidth]{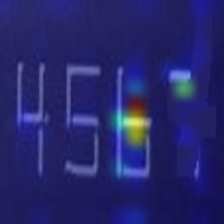}& \includegraphics[width=0.15\linewidth, height=0.15\linewidth]{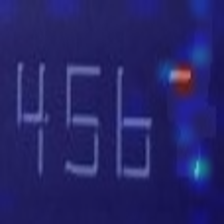}\\ &\raisebox{3\normalbaselineskip}[0pt][0pt]{\rotatebox[origin=c]{90}{\small  \tb{DIFF.}}}&\includegraphics[width=0.15\linewidth,  height=0.15\linewidth]{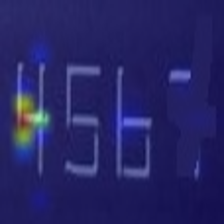}& \includegraphics[width=0.15\linewidth, height=0.15\linewidth]{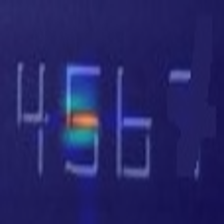}& \includegraphics[width=0.15\linewidth, height=0.15\linewidth]{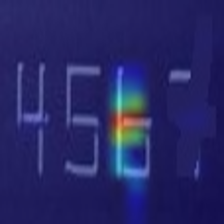} & \includegraphics[width=0.15\linewidth, height=0.15\linewidth]{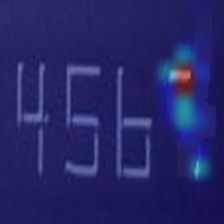}\\[10ex]

\multirow{2}{*}[0.92in]{\includegraphics[width= 0.3\linewidth, height=0.31\linewidth]{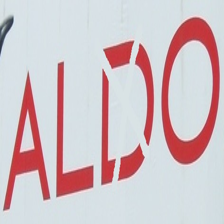}} &  \raisebox{3\normalbaselineskip}[0pt][0pt]{\rotatebox[origin=c]{90}{\small \tb{STANDARD}}} & \includegraphics[width=0.15\linewidth, height=0.15\linewidth]{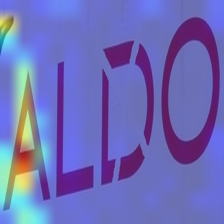}& \includegraphics[width=0.15\linewidth, height=0.15\linewidth]{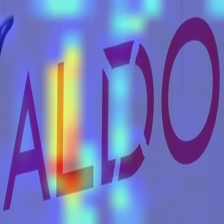}& \includegraphics[width=0.15\linewidth, height=0.15\linewidth]{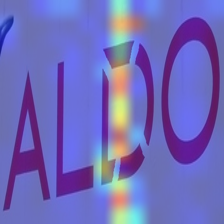}& \includegraphics[width=0.15\linewidth, height=0.15\linewidth]{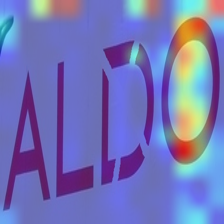}\\ &\raisebox{3\normalbaselineskip}[0pt][0pt]{\rotatebox[origin=c]{90}{\small  \tb{DIFF.}}}&\includegraphics[width=0.15\linewidth,  height=0.15\linewidth]{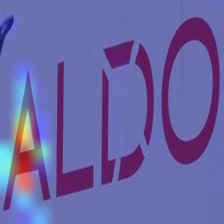}& \includegraphics[width=0.15\linewidth, height=0.15\linewidth]{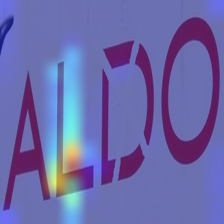}& \includegraphics[width=0.15\linewidth, height=0.15\linewidth]{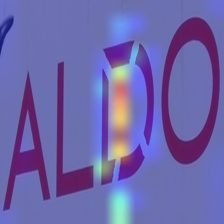} & \includegraphics[width=0.15\linewidth, height=0.15\linewidth]{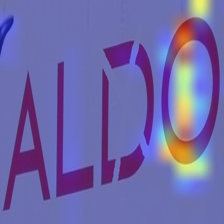}\\[10ex]

\multirow{2}{*}[0.92in]{\includegraphics[width= 0.3\linewidth, height=0.31\linewidth]{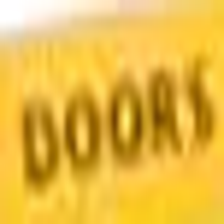}} &  \raisebox{3\normalbaselineskip}[0pt][0pt]{\rotatebox[origin=c]{90}{\small \tb{STANDARD}}} & \includegraphics[width=0.15\linewidth, height=0.15\linewidth]{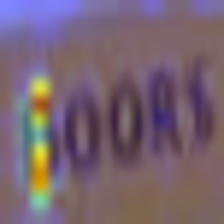}& \includegraphics[width=0.15\linewidth, height=0.15\linewidth]{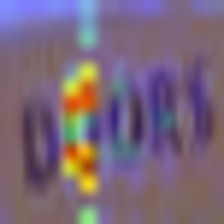}& \includegraphics[width=0.15\linewidth, height=0.15\linewidth]{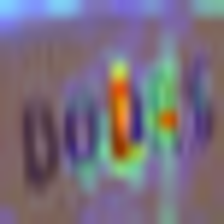}& \includegraphics[width=0.15\linewidth, height=0.15\linewidth]{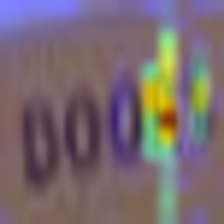}\\ &\raisebox{3\normalbaselineskip}[0pt][0pt]{\rotatebox[origin=c]{90}{\small  \tb{DIFF.}}}&\includegraphics[width=0.15\linewidth,  height=0.15\linewidth]{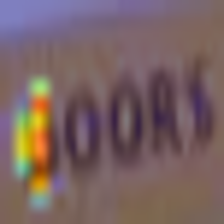}& \includegraphics[width=0.15\linewidth, height=0.15\linewidth]{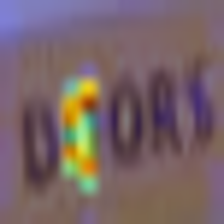}& \includegraphics[width=0.15\linewidth, height=0.15\linewidth]{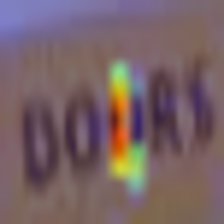} & \includegraphics[width=0.15\linewidth, height=0.15\linewidth]{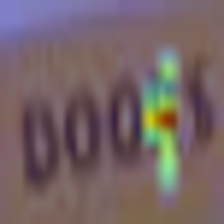}\\[10ex]
\end{tabular}}
\caption{\tb{Comparison of Attention Maps.} Attention maps of the last Cross-Attention in the last block of the Permutation Language Decoder. On the left: the original input image. First row of each section: attention maps obtained with the standard Cross-Attention. Second row of each section: attention maps obtained with our Differential Cross-Attention.}
\label{tab:maps3}
\end{table*}
In Sec.~\ref{subsec:diff}, we presented our Permutation Language Decoder equipped with Differential Cross-Attention layers. The aim was to minimize the amount of noise present in the attention maps. Tables~\ref{tab:maps1}, \ref{tab:maps2} and \ref{tab:maps3}  provide visual comparisons between the standard Cross-Attention and our Differential Cross-Attention. From the results, the majority of the noise and errors observed in the standard Cross-Attention are effectively reduced when the Differential Cross-Attention is used.


\section{Effectiveness of CSD}
\label{sec:additional_considerations}
All the components of CSD, namely pseudo-labels, knowledge distillation of the context-aware predictions and differential decoder, provide substantial improvements as
presented in Tab.~\ref{tab:csd_results_pkd} and Tab.~\ref{tab:data_scaling}. Specifically, on Real dataset with the 
base model, pseudo-labels (PL) provide $+0.26\%$ improvement by themselves. When PL and context-aware KD are combined, the improvement is $+0.50\%$ (providing robustness to label noise). Finally, the differential decoder (DD) provides an additional relevant improvement (mitigating attention noise): $\text{PL}+\text{KD}+\text{DD}$ obtains $+0.70\%$. Notably, many benchmarks in STR (used to compute the average accuracy) are saturated and affected by test label errors. For this reason, while the improvements might seem modest, they are significant. For comparison, CLIP4STR scales the architecture from 158M to 446M parameters to obtain only $+0.56\%$ improvement.

\section{Additional Results}
\label{sec:qualitative_examples}
In Table \ref{tab:qual1} and \ref{tab:qual2}, we present qualitative examples of predictions of our STR model by comparing with CLIP4STR \cite{zhao2023clip4str}. From the results, even if CLIP4STR has a separate branch for text correction, our STR model obtains more accurate results, especially for occluded cases. This shows that training encoder-decoder parts together provides robustness and improves the accuracy. In Table \ref{tab:union_results} we show that CSD outperforms previous state-of-the-art even in the challenging Union14M benchmark.

\begin{table*}
  \centering
  \resizebox{0.99\textwidth}{!}{
  \begin{tabular}{ll}
  \begin{tabular}{cccc}
   \textbf{Image} &  \textbf{Ground Truth}  & \tb{CLIP4STR-L} & \tb{CSD-D (ours)} \\[2ex]
   \makecell{\includegraphics[width=0.2\textwidth, height=0.2\textwidth]{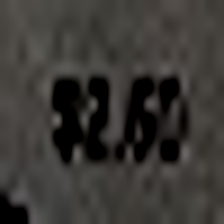}} & \texttt{260} & \texttt{2\textcolor{red}{5}0} & \texttt{260} \\
   \makecell{\includegraphics[width=0.2\textwidth, height=0.2\textwidth]{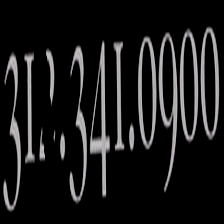}} & \texttt{3123410900} & \texttt{31\textcolor{red}{1}3410900} & \texttt{3123410900} \\
    \makecell{\includegraphics[width=0.2\textwidth, height=0.2\textwidth]{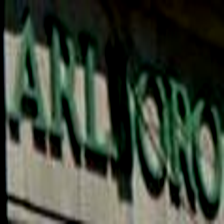}} & \texttt{arlboro} & \texttt{arl\textcolor{red}{j}oro} & \texttt{arl\textcolor{red}{i}oro} \\
    \makecell{\includegraphics[width=0.2\textwidth, height=0.2\textwidth]{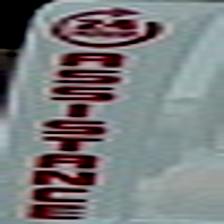}} & \texttt{assistence} & \texttt{\textcolor{red}{w}assistence} & \texttt{assistence} \\
    \makecell{\includegraphics[width=0.2\textwidth, height=0.2\textwidth]{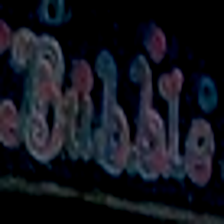}} & \texttt{bubble} & \texttt{bubble} & \texttt{b\textcolor{red}{i}bble} \\
    \makecell{\includegraphics[width=0.2\textwidth, height=0.2\textwidth]{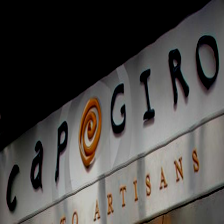}} & \texttt{capogiro} & \texttt{cap\textcolor{red}{\_}giro} & \texttt{capogiro} \\
    \makecell{\includegraphics[width=0.2\textwidth, height=0.2\textwidth]{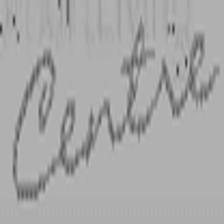}} & \texttt{centre} & \texttt{cent\textcolor{red}{i}e} & \texttt{centre} \\
  \end{tabular}
  & 

  \begin{tabular}{cccc}
   \textbf{Image} &  \textbf{Ground Truth}  & \tb{CLIP4STR-L} & \tb{CSD-D (ours)} \\[2ex]
   
   \makecell{\includegraphics[width=0.2\textwidth, height=0.2\textwidth]{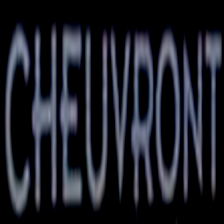}} & \texttt{cheuvront} & \texttt{cheu\textcolor{red}{\_}ront} & \texttt{cheuvront} \\
    \makecell{\includegraphics[width=0.2\textwidth, height=0.2\textwidth]{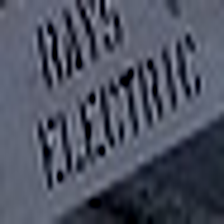}} & \texttt{electric} & \texttt{elect\textcolor{red}{n}ic} & \texttt{electric} \\
    \makecell{\includegraphics[width=0.2\textwidth, height=0.2\textwidth]{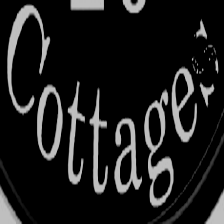}} & \texttt{cottages} & \texttt{cottages} & \texttt{cottage\textcolor{red}{e}} \\
    \makecell{\includegraphics[width=0.2\textwidth, height=0.2\textwidth]{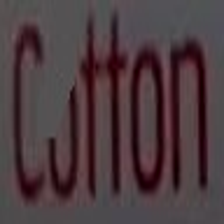}} & \texttt{cotton} & \texttt{c\textcolor{red}{u}tton} & \texttt{c\textcolor{red}{u}tton} \\
    \makecell{\includegraphics[width=0.2\textwidth, height=0.2\textwidth]{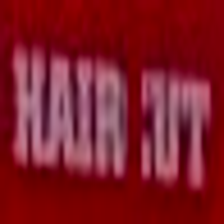}} & \texttt{haircut} & \texttt{hai\textcolor{red}{\_}cut} & \texttt{haircut} \\
    \makecell{\includegraphics[width=0.2\textwidth, height=0.2\textwidth]{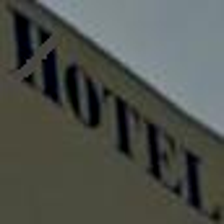}} & \texttt{hotel} & \texttt{\textcolor{red}{l}otel} & \texttt{hotel} \\
    \makecell{\includegraphics[width=0.2\textwidth, height=0.2\textwidth]{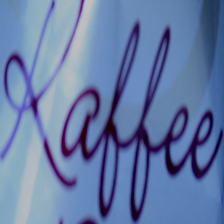}} & \texttt{kaffee} & \texttt{\textcolor{red}{l}affee} & \texttt{kaffee} \\
  \end{tabular}
  \end{tabular}}
  
  \caption{\tb{Qualitative examples.} The table presents image examples along with ground truth labels, predictions made by our models, CSD-D and CLIP4STR-L, which were both trained using the RBU dataset. These predictions are based on a character set consisting of 36 alphanumeric characters. Errors are highlighted in \textcolor{red}{red}.}
   \label{tab:qual1}
  \vspace{-1.2em}
\end{table*}

\begin{table*}
  \centering
  \resizebox{0.99\textwidth}{!}{
  
  \begin{tabular}{ll}
  \begin{tabular}{cccc}
   \textbf{Image} &  \textbf{Ground Truth}  & \tb{CLIP4STR-L} & \tb{CSD-D (ours)} \\[2ex]
   
   \makecell{\includegraphics[width=0.2\textwidth, height=0.2\textwidth]{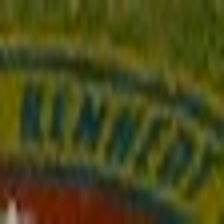}} & \texttt{kennedy} & \texttt{kenned\textcolor{red}{\_}} & \texttt{kennedy} \\
    \makecell{\includegraphics[width=0.2\textwidth, height=0.2\textwidth]{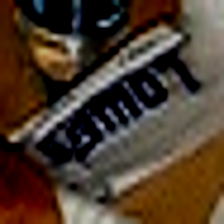}} & \texttt{lower} & \texttt{\textcolor{red}{p}ower} & \texttt{\textcolor{red}{\_}ower} \\
     \makecell{\includegraphics[width=0.2\textwidth, height=0.2\textwidth]{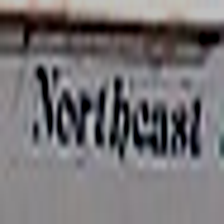}} & \texttt{northeast} & \texttt{n\textcolor{red}{e}rtheast} & \texttt{northeast} \\
      \makecell{\includegraphics[width=0.2\textwidth, height=0.2\textwidth]{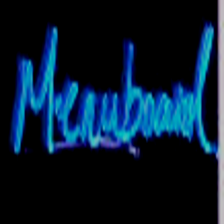}} & \texttt{menuboard} & \texttt{menuboard} & \texttt{me\textcolor{red}{r}uboard} \\
      \makecell{\includegraphics[width=0.2\textwidth, height=0.2\textwidth]{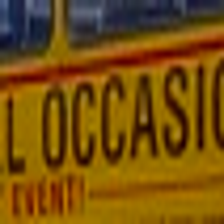}} & \texttt{\textcolor{red}{l}occasio} & \texttt{\textcolor{red}{l}occasi\textcolor{red}{o}} & \texttt{loccasio} \\
      \makecell{\includegraphics[width=0.2\textwidth, height=0.2\textwidth]{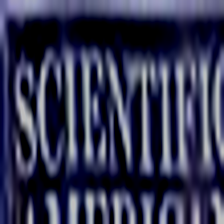}} & \texttt{scientific} & \texttt{scientifi\textcolor{red}{\_}} & \texttt{scentifi\textcolor{red}{\_}} \\
      \makecell{\includegraphics[width=0.2\textwidth, height=0.2\textwidth]{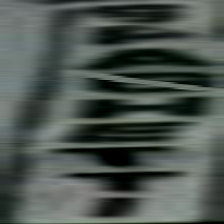}} & \texttt{spaghetti} & \texttt{spa\textcolor{red}{c}hetti} & \texttt{spaghetti} \\
  \end{tabular}

  &

  \begin{tabular}{cccc}
   \textbf{Image} &  \textbf{Ground Truth}  & \tb{CLIP4STR-L} & \tb{CSD-D (ours)} \\[2ex]
   
   \makecell{\includegraphics[width=0.2\textwidth, height=0.2\textwidth]{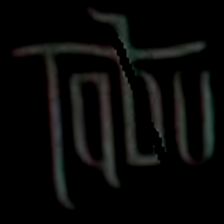}} & \texttt{tabu} & \texttt{t\textcolor{red}{q}bu} & \texttt{tabu} \\
    \makecell{\includegraphics[width=0.2\textwidth, height=0.2\textwidth]{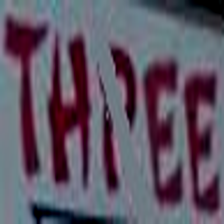}} & \texttt{three} & \texttt{th\textcolor{red}{p}ee} & \texttt{three} \\
    \makecell{\includegraphics[width=0.2\textwidth, height=0.2\textwidth]{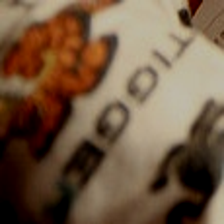}} & \texttt{tigger} & \texttt{tigge\textcolor{red}{n}} & \texttt{tigger} \\
    \makecell{\includegraphics[width=0.2\textwidth, height=0.2\textwidth]{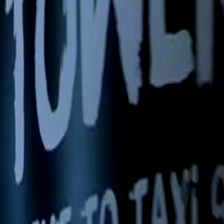}} & \texttt{towe\textcolor{red}{\_}} & \texttt{tower} & \texttt{tower} \\
    \makecell{\includegraphics[width=0.2\textwidth, height=0.2\textwidth]{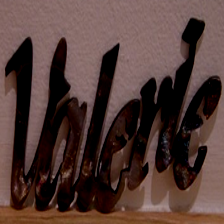}} & \texttt{valerie} & \texttt{valer\textcolor{red}{t}e} & \texttt{valerie} \\
    \makecell{\includegraphics[width=0.2\textwidth, height=0.2\textwidth]{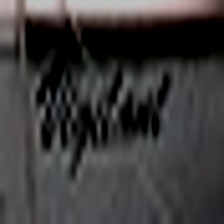}} & \texttt{vigilant} & \texttt{vigi\textcolor{red}{t}ant} & \texttt{vigi\textcolor{red}{t}ant} \\
     \makecell{\includegraphics[width=0.2\textwidth, height=0.2\textwidth]{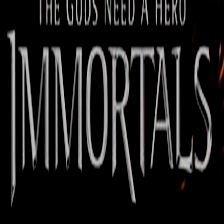}} & \texttt{immortals} & \texttt{immortals} & \texttt{\textcolor{red}{j}mmortals} \\
  \end{tabular}
  \end{tabular}
  }
  
  \caption{\tb{Qualitative examples.} The table presents image examples along with ground truth labels, predictions made by our models, CSD-D and CLIP4STR-L, which were both trained using the RBU dataset. These predictions are based on a character set consisting of 36 alphanumeric characters. Errors are highlighted in \textcolor{red}{red}.}
  \label{tab:qual2}
  \vspace{-1.2em}
\end{table*}

\begin{table*}[t!]
  \centering
  \resizebox{0.9\linewidth}{!}{
  \addtolength{\tabcolsep}{-0.3em}
  \begin{tabular}{lllcccccccc}
    \toprule
     \textbf{Method} &  \textbf{Data} & \textbf{Params}  & \textbf{Curve} & \textbf{Multi-Oriented} & \textbf{Artistic} & \textbf{Contextless} & \textbf{Salient} & \textbf{Multi-Words} & \textbf{General} & \textbf{Avg}\\
    \midrule

     CLIP4STR-B & Real & 158M  & 96.3 & 96.1 & 86.5 & \textbf{92.2} & 91.2 & 88.9 & 89.9 & 91.6 \\
    CLIP4STR-L & Real & 446M        & \textbf{97.0} & 96.6 & 87.2 & 91.0 & 91.5 & \textbf{89.9} & 90.3 & 91.9 \\
    \textbf{CSD-D (ours)} & Real & \textbf{110M} & \textbf{97.0} & \textbf{97.0} & \textbf{87.7} & 91.8 & \textbf{91.7} & 89.5 & \textbf{91.7} & \textbf{92.3} \\
    
    \midrule
    CLIP4STR-B & REBU-Syn & 158M & 96.4 & 96.3 & \textbf{88.6} & 90.1 & 91.9 & \textbf{92.2} & 89.1 & 92.1 \\
    CLIP4STR-L & REBU-Syn & 446M & 96.4 & \textbf{97.2} & \textbf{88.6} & 90.4 & 92.7 & 90.7 & 89.3 & 92.2 \\
    \textbf{CSD-D (ours)} & RBU & \textbf{110M} & \textbf{96.5} & \textbf{97.2} & \textbf{88.6} & \textbf{92.8} & \textbf{92.8} & 90.8 & \textbf{90.2} & \textbf{92.7} \\
    \bottomrule
 \end{tabular}}
  
  \vspace{-0.8em}
  \caption{Comparison of CSD-D with CLIP4STR (results from \cite{rang2023large}) on Union14M benchmark.}
  \label{tab:union_results}
  \vspace{-2.0em}
\end{table*}


\end{document}